\documentclass[lettersize,journal]{IEEEtran}
\usepackage{amsmath,amsfonts,amssymb}
\usepackage{algorithmic}
\usepackage{algorithm}
\usepackage{array}
\usepackage{textcomp}
\usepackage{color}
\usepackage{stfloats}
\usepackage{url}
\usepackage{verbatim}
\usepackage{graphicx}
\usepackage{cite}
\usepackage{graphicx}
\usepackage{amsmath}
\usepackage{amssymb}
\usepackage{multirow}
\usepackage{tabularx}
\usepackage{booktabs}
\usepackage{diagbox}
\usepackage{paralist}
\usepackage{epstopdf}
\usepackage{color}
\usepackage{subfig}
\usepackage{multirow}
\usepackage{multicol}
\usepackage{color}
\usepackage{xcolor}
\usepackage{verbatim}
 
\newcommand{\para}[1]{\noindent \textbf{#1}}

\begin{document}

\title{Trustworthy Multimodal Fusion for Sentiment Analysis in Ordinal Sentiment Space}

\author{Zhuyang Xie,
		Yan Yang*,~\IEEEmembership{Member,~IEEE}, 
		Jie Wang,
		Xiaorong Liu,
		and Xiaofan Li
\thanks{Zhuyang Xie, Yan Yang, Jie Wang, Xiaorong Liu, and Xiaofan Li are with the School of Computing and Artificial Intelligence, Southwest Jiaotong University, Chengdu, 999 Xi’an Rd, 611756, China, and also with Manufacturing Industry Chains Collaboration and Information Support Technology Key Laboratory of Sichuan Province, Southwest Jiaotong University, Chengdu 611756, China. E-mail: zyxie@my.swjtu.edu.cn, yyang@swjtu.edu.cn, JackWang@my.swjtu.edu.cn, xr\_liu@my.swjtu.edu.cn, wonk\_lixiaofan@163.com.}
\thanks{* is the corresponding author.}
}



\maketitle
\makeatletter

\IEEEpubid{\begin{minipage}{\textwidth} \ \\[35pt] \centering
		Copyright~\copyright~2024 IEEE. Personal use of this material is permitted. However, permission to use this material for any other purposes \\ must be obtained from the IEEE by sending an email to pubs-permissions@ieee.org.
	\end{minipage}
}
\begin{abstract}
Multimodal video sentiment analysis aims to integrate multiple modal information to analyze the opinions and attitudes of speakers. Most previous work focuses on exploring the semantic interactions of intra- and inter-modality. However, these works ignore the reliability of multimodality, i.e., modalities tend to contain noise, semantic ambiguity, missing modalities, etc. In addition, previous multimodal approaches treat different modalities equally, largely ignoring their different contributions. Furthermore, existing multimodal sentiment analysis methods directly regress sentiment scores without considering ordinal relationships within sentiment categories, with limited performance. To address the aforementioned problems, we propose a trustworthy multimodal sentiment ordinal network (TMSON) to improve performance in sentiment analysis. Specifically, we first devise a unimodal feature extractor for each modality to obtain modality-specific features. Then, an uncertainty distribution estimation network is customized, which estimates the unimodal uncertainty distributions. Next, Bayesian fusion is performed on the learned unimodal distributions to obtain multimodal distributions for sentiment prediction. Finally, an ordinal-aware sentiment space is constructed, where ordinal regression is used to constrain the multimodal distributions. Our proposed TMSON outperforms baselines on multimodal sentiment analysis tasks, and empirical results demonstrate that TMSON is capable of reducing uncertainty to obtain more robust predictions.
\end{abstract}


\section{Introduction}\label{sec:intro}
\IEEEPARstart{W}{ith} the daily activities of humans in social media, massive personal videos are available, and it is particularly important to analyze these multimodal videos. Multimodal sentiment analysis aims to integrate multimodal information (including text, visual, and audio) into a consistent representation. Combining multiple modalities can provide complementary information to facilitate emotion recognition, which poses two challenges. The first challenge is to explore intramodal dynamics and intermodal dynamics, where intramodal dynamics refer to the changes within unimodal interactions, and intermodal dynamics refer to cross-modal interactions. Another challenge is that multimodal sequences are typically misaligned and heterogeneous, which hinders multimodal fusion.

To address the first challenge, some works employ RNN-based methods for temporal modeling~\cite{zadeh2017tensor,hazarika2020misa,pham2019found,sun2020learning}. However, these methods focus only on intramodal dynamics and ignore intermodal dynamics. A workaround is to introduce extra memory for cross-modal interactions~\cite{zadeh2018memory}. In addition, attention-based approaches are also used to explore cross-modal dynamics~\cite{hasan2021humor,gu2018multimodal,zadeh2018multi}. To address the second challenge, one solution is to use an encoder-decoder structure to enable translation between heterogeneous modalities~\cite{pham2019found,pham2018seq2seq2sentiment}. The other is to minimize the heterogeneity gap between modalities in a common feature space~\cite{hazarika2020misa}.

\begin{figure}[t]
	\centering
	\includegraphics[scale = 0.48]{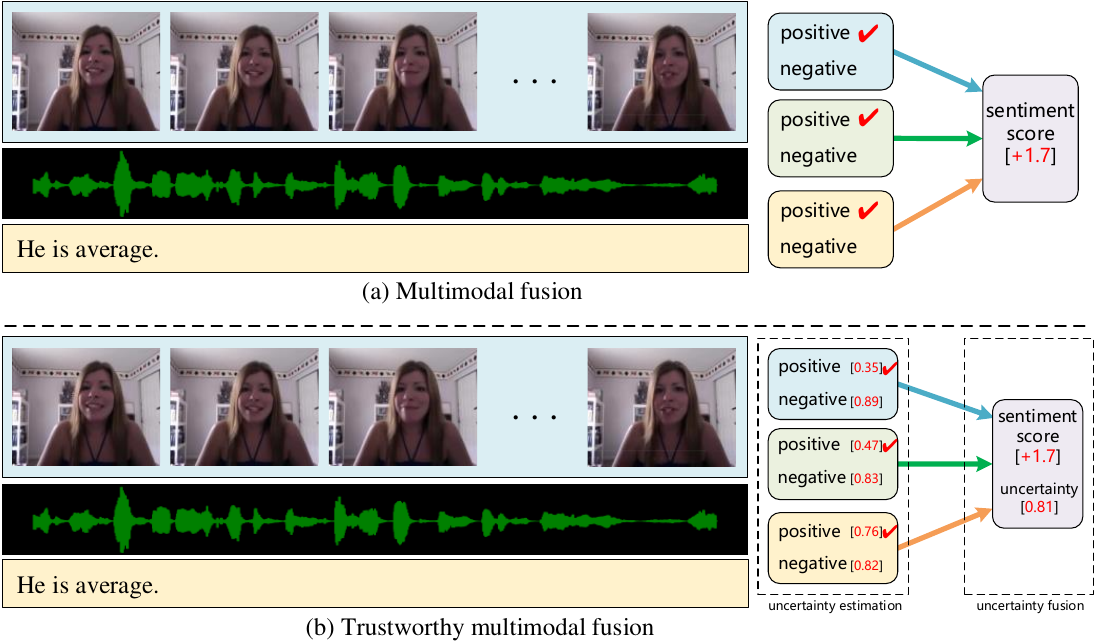}
	\caption{Illustration of the differences between previous multimodal fusion methods and our trustworthy multimodal fusion method. (a) The aim of multimodal fusion is to integrate different modal information into a consistent representation for sentiment analysis without providing reliability for the prediction results. (b) Trustworthy multimodal fusion, in contrast, estimates the uncertainty of different modalities and uses uncertainty distribution fusion for more robust sentiment prediction. The numbers in brackets indicate uncertainty scores, with larger values indicating higher uncertainty.}
	\label{motivation}
\end{figure}

However, the abovementioned methods do not perform multimodal analysis from a reliability perspective. Previous multimodal methods directly extract modality-specific features based on deterministic embeddings and integrate trimodal information to predict sentiment scores $\bigl($shown in Fig.~\ref{motivation}(a)$\bigr)$. However, these modalities tend to contain noise, semantic ambiguity, etc., which hinder sentiment prediction and may not be trustworthy. In contrast, our method provides uncertainty estimates for the observed unimodal features and uses distribution fusion to obtain multimodal representations. Specifically, our uncertainty distribution estimation module computes a sentiment score for each modality and predicts its uncertainty, with the uncertainty score reflecting the confidence of each modality. Fig.~\ref{motivation}(b) shows the case of semantic ambiguity. If only visual and audio modalities are considered, a positive classification with a high confidence score is obtained. However, when confronted with text modality, a low-confidence classification is obtained. The reason is that \emph{``He is average"} in the video clip does not reflect the speaker’s attitude within the full context: \emph{``He is average, and its sh so short lived, \ldots\ldots His fight sequences are very neat, and he delivers a lot of intensity"} expresses positive comments. Although our final forecast is positive, it has a high uncertainty score, which reflects how trustworthy the decision is.

In recent years, multimodal machine learning has emerged in various fields, benefiting from the complementary information provided by each modality, which further improves the model performance. However, multimodal data may contain noise during acquisition and transmission, and the intensity of content conveyed by each modality varies. At this point, we need to consider the reliability of these multiple modalities, i.e., how to weigh modalities of different qualities and whether our model’s predictions can be trusted. Most existing feature extraction approaches use deep neural networks to perform feature transformation and obtain deterministic point representations in the feature space. Although these methods work well, they require clean or high-quality data. When working with low-quality data, ambiguous sample embeddings may cause shifts, leading to large errors. Some uncertain embedding methods have been proposed to solve this problem~\cite{vilnis2015word,oh2018modeling,shi2019probabilistic}. Vilnis et al~\cite{vilnis2015word} adopt a probabilistic embedding method to represent the uncertainty of word embeddings in a Gaussian distribution space. HIB~\cite{oh2018modeling} employs hedge instance embeddings, which are modeled as random variables to represent input uncertainty. PFE~\cite{shi2019probabilistic} uses the Gaussian distribution to model the uncertainty of face images, where the mean represents the feature values and the variance is used to measure the uncertainty. Several recent works model uncertainty by estimating the distribution of observed features. TMC~\cite{han2021trusted} proposes a multiview classification method that incorporates the Dempster-Shafer theory to integrate different views at the evidence level and improves model robustness. DUA-Nets~\cite{geng2021uncertainty} dynamically performs uncertainty estimation according to the data quality of different views under unsupervised conditions and uses uncertainty estimation to exploit high-quality view data.

We propose TMSON to dynamically estimate uncertainty for multimodal data of varying quality for robust multimodal fusion. Motivated by DUA-Nets~\cite{geng2021uncertainty}, we hypothesize that the observed data samples are generated from Gaussian distributions, and the distribution of each modal also varies. Based on the observed modalities, an uncertainty distribution estimation module is customized, which estimates the unimodal distributions, where the mean values of the distribution represent the modal identity and the variance represents the uncertainty. Different from previous work on multimodal sentiment analysis, we propose a multimodal distribution fusion module to achieve multimodal fusion. Our distribution fusion mechanism uses Bayes' rule, and conceptually, the fused multimodal distribution has a more robust mean and smaller variance. The experiments conducted demonstrate that the uncertainty is effectively reduced from unimodal to multimodal during the fusion process. To incorporate the ordinal relationship of sentiment categories into the model, ordinal regression is introduced to constrain the sentiment space so that the fused multimodal distribution is forced to follow the sentiment order. Furthermore, we explore multitask learning, jointly training unimodal and multimodal tasks to learn modality-specific and modality-shared representations. Extensive experiments also show that TMSON can capture uncertainty and achieves superior performance compared to baselines in the context of noise disturbance.

The main contributions of this paper are as follows:
\begin{itemize}
	\item A trustworthy multimodal sentiment analysis model is proposed. The uncertainty distribution estimation module dynamically estimates the uncertainty distribution for each modality. Furthermore, a multimodal distribution fusion module is introduced to fuse unimodal distributions. The fused multimodal distribution has smaller variance and more robust performance. 
	\item To inject the ordinal relationships within sentiment categories into the model, an ordinal regression loss is introduced in the sentiment space to constrain the multimodal distributions. Experiments show that under the ordinal constraint, the misclassified samples are closer to their actual class.
	\item Extensive experiments show that TMSON outperforms baselines on sentiment analysis tasks. Moreover, our model maintains superior performance in terms of stability.
\end{itemize}

\section{Related Work}
\subsection{Multimodal Sentiment Analysis}
In recent years, multimodal sentiment analysis, which integrates heterogeneous data (e.g., text, vision, audio) in video mainly by exploring the consistency and complementarity of multiple modalities, has attracted researchers’ attention. Previous research focused mainly on multimodal representation and multimodal fusion.

For multimodal representations, MISA~\cite{hazarika2020misa} projects each modality into two subspaces to learn modality-invariant and modality-specific features. MIB~\cite{mai2022multimodal} introduces an information bottleneck to learn multimodal representations by maximizing the mutual information (MI) between the representation and the target while constraining the MI between the representation and the input data. Self-MM~\cite{yu2021learning} combines self-supervised methods to design a unimodal label generation module, which learns informative unimodal representations by simultaneously learning multimodal and unimodal tasks. RAVEN~\cite{wang2019words} considers the shift in language caused by nonverbal environments, models nonverbal subword sequences and dynamically transforms word representations based on nonverbal cues. 

For multimodal fusion, there is considerable literature on feature-level fusion. The two direct methods of feature-level fusion are early fusion and late fusion. A direct fusion method for early fusion is concatenation, which fuses unimodal features at the input level~\cite{rozgic2012ensemble,poria2017context}. This method is a simple input-level concatenation and is not sufficient to explore cross modal interactions. Late fusion tends to integrate unimodal decisions through voting, weighting, etc., to make the final decision~\cite{zhang2017learning,nojavanasghari2016deep,kampman2018investigating}. In addition, there are some multimodal fusion methods, such as  graph-based fusion methods~\cite{chu2014effective,yang2020mtgat,mai2020modality,zadeh2018multimodal}, tensor-based fusion methods~\cite{zadeh2017tensor,liu2018efficient}, and attention-based fusion methods~\cite{ou2020multimodal,hazarika2020misa,wang2022m3s}. Some recent works focus on feature fusion over time steps, involving unaligned multimodal sequence fusion~\cite{wu2021graph,liang2021attention,lv2021progressive,mai2020multi}. 

We focus on unaligned multimodal sequences and differ from previous work that designs sophisticated networks to model crossmodal relationships. In the fusion stage, our multimodal distribution fusion module directly takes the unimodal distributions as input without additional learnable parameters. Furthermore, TMSON is trained on unimodal and multimodal tasks simultaneously, thereby learning modality-specific and modality-shared features. 

\subsection{Trustworthy Multimodal Learning}
In practice, various sensors or environmental factors can affect the quality of each modality. When analyzing these multimodal data, low-quality modalities will seriously hinder multimodal tasks. Therefore, estimating uncertainty and making trustworthy decisions are critical.

There are two main types of uncertainty: data uncertainty and model uncertainty. Data uncertainty refers to noise inherent in training data that cannot be alleviated with more data. Model uncertainty, which is related to the model parameters, is caused by incomplete training or data. Recently, several studies~\cite{han2021trusted,geng2021uncertainty} have focused on uncertainty estimation in deep learning to obtain more robust performance and provide interpretability. Uncertainty estimation has attracted extensive attention in the field of computer vision, such as age estimation~\cite{shi2019probabilistic,chang2020data}, semantic segmentation~\cite{huang2018efficient,rottmann2019uncertainty}, and object detection~\cite{kraus2019uncertainty,harakeh2020bayesod}. Some methods employ stochastic embeddings to model data uncertainty~\cite{oh2018modeling,shi2019probabilistic}. Other approaches use probabilistic models to model data uncertainty to reduce the effects of noise~\cite{kendall2017uncertainties,amini2020deep}. Coincidentally, reliability research has also emerged in the multiview/multimodal field~\cite{han2021trusted,han2022trusted,han2022multimodal,ma2021trustworthy}. Some methods make uncertain estimates of observational evidence at the feature level, resulting in reliable predictions~\cite{han2021trusted,han2022trusted}. MoNIG~\cite{ma2021trustworthy} is an extension of work~\cite{amini2020deep}, using a mixture of normal-inverse gamma distributions to estimate the uncertainty of different modalities and produce reliable regression results. Multimodal dynamics~\cite{han2022multimodal} automatically evaluates the informativeness for different samples at the feature level and modality level, enabling reliable integration of multiple modalities.

Our work focuses on data uncertainty in multimodal sentiment analysis, directly observing each modality and extracting modality-specific features that are used to estimate the uncertainty distributions. We design an uncertainty distribution estimation module that uses distributional representations instead of deterministic embeddings to automatically estimate the importance of different modalities for different samples.

\subsection{Ordinal Regression}
Many realistic labels follow a natural order with each other. Taking the CMU-MOSI~\cite{zadeh2016mosi} dataset as an example, we can refine the sentiment score into 7 levels: \{\emph{highly negative, negative, weakly negative, neutral, weakly positive, positive, highly positive}\}. Therefore, direct regression, trained with a single output neuron to predict the target value, ultimately has limited performance. There are two main reasons. First, the cost of different misclassifications is different; the more mistakes there are, the greater the penalty is. Second, introducing ranking information and learning interclass ordinal relationships help to build more accurate models.

Ordinal regression problems are explored in many domains, such as age estimation~\cite{niu2016ordinal,li2021learning,shin2022moving,li2022unimodal}, image segmentation~\cite{zhao2019ordinal,diaz2019soft}, and depth estimation~\cite{fu2018deep,herrmann2020learning,kreuzig2019distancenet}. AFAD~\cite{niu2016ordinal} uses a multi-output CNN to transform ordinal regression into multiple binary classification subproblems. POE~\cite{li2021learning} employs probabilistic ordinal embedding and models the ordinal relationship of age by constraining the ordinal distribution. MWR~\cite{shin2022moving} introduces the concept of relative rank ($\rho$-rank) in the ordinal relation, which is used to represent the ordinal relationship of input and reference instances. DORN~\cite{fu2018deep} adopts a discretization strategy to discretize depth and redefines depth estimation as ordinal regression.

Previous work has approached sentiment analysis using direct regression without considering the natural ordering between sentiment categories. In the multimodal sentiment space, an ordinal constraint is introduced to learn the ordinal dependencies between different emotion categories. With ordinal regression, the performance of sentiment analysis is further improved.

\section{Methodology}\label{Sec:method}
\begin{figure*}[ht]
	\centering
	\includegraphics[width=\textwidth, height=9cm]{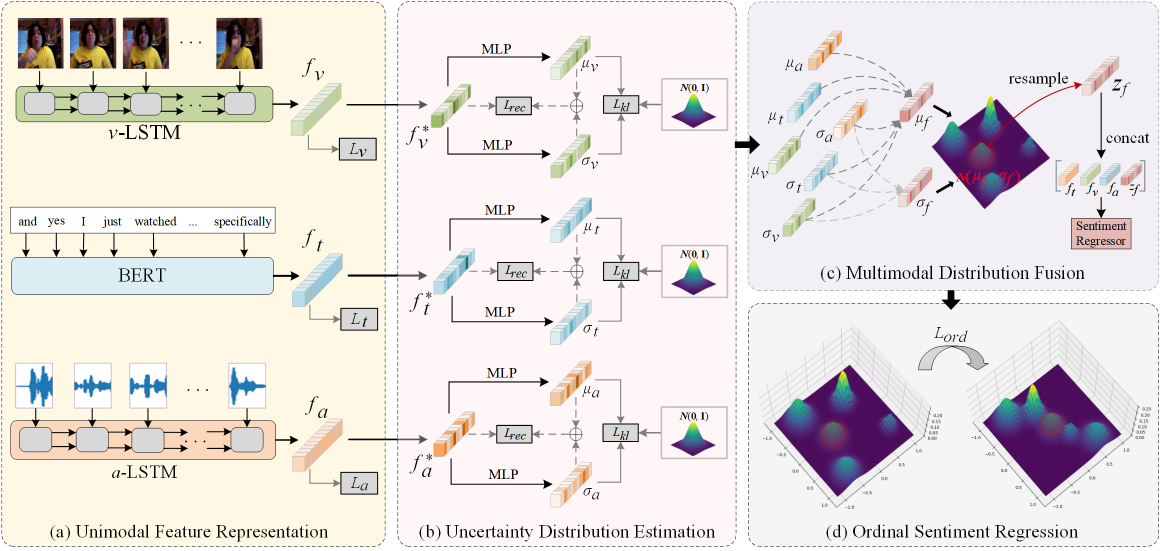}
	\caption{The diagram of the TMSON framework. The input to the network is multimodal sequences (text, visual, and audio). For each modality, we observe unimodal representation (a) and estimate the uncertainty distribution (b), after which we fuse these distributions to obtain a consistent multimodal distribution (c). Ultimately, ordinal regression is introduced to constrain the fused multimodal distribution to be ordinal (d).}
	\label{model}
\end{figure*}

Previous work has focused on developing fusion strategies based on exploring cross-modal interactions, while ignoring the quality and information contributions of different modalities. Innovated by trusted multimodal learning, we propose TMSON, which adaptively estimates Gaussian distributions for various modalities, where the mean vector represents modal intensity and the variance measures the uncertainty. For unimodal distributions, TMSON naturally applies Bayes' rule for fusion, akin to the Kalman filter. Afterward, we constrain the fused multimodal distributions in ordinal space, enabling the model to perform ordinal-aware emotion recognition.

The proposed pipeline is described in detail as shown in Fig.~\ref{model}. We first elaborate on the problem and its notation and then introduce the proposed TMSON, which is detailed in four subsections, namely, unimodal feature representation, uncertainty distribution estimation, multimodal distribution fusion, and ordinal sentiment regression. Finally, training and optimization objectives are introduced.

\subsection{Problem Formulation}
Our goal is to predict sentiment scores using multimodal data to facilitate  sentiment analysis. The input is an utterance $U_m \in \mathbb{R}^{T_m \times d_m}$, where $m \in \{t,v,a\}$,  $T_m$ represents the length of  sequence $m$, and $d$ represents the feature dimension. Each utterance contains three sequences: text ($t$), visual ($v$), and audio ($a$). These sequences are represented as $U_{t} \in \mathbb{R}^{T_t \times d_t}$, $U_{v} \in \mathbb{R}^{T_v \times d_v}$, and $U_{a} \in \mathbb{R}^{T_a \times d_a}$, respectively. Given an utterance $U_{m}$, we formulate the multimodal sentiment analysis task as follows:
\begin{equation}
\label{eqn_1}
f_m = E_m(U_m; \,\theta_m), \; m \in \{t,v,a\}
\end{equation}
\begin{equation}
\label{eqn_2}
\hat{y}_f = FC\bigl(Fusion(f_t, f_v, f_a); \, \theta_f \bigr)
\end{equation}
where $E_m$ is the unimodal feature extractor and is parameterized by $\theta_m$; $f_m$ is the unimodal feature observed by $E_m$; $Fusion(\cdot)$ is the multimodal fusion strategy; and $FC$ is a fully connected layer, which is parameterized by $\theta_f$ to predict the multimodal sentiment score $\hat{y}_f$. The parameters $\theta$ of the overall model are optimized as follows:
\begin{equation}
\label{eqn_3}
\mathcal{L}_{f} = |\hat{y}_f - y|
\end{equation}
\begin{equation}
\label{eqn_4}
\theta {\longleftarrow} \, \theta - \alpha \frac{\partial \mathcal{L}_{f}}{\partial \theta}
\end{equation}
where $y$ is the ground truth label, $|\cdot|$ is the mean absolute error (MAE), $\mathcal{L}_{f}$ is the multimodal loss, and $\alpha$ is the learning rate.

We explore multitask learning to decouple unimodal and multimodal learning processes---learning modality-specific and modality-shared features. Specifically, we take the unimodal feature $f_m$ obtained from Eq.~(\ref{eqn_1}) for unimodal loss computation $\mathcal{L}_{m}$, learning modality-specific features. Combining multimodal loss $\mathcal{L}_{f}$ with unimodal loss $\mathcal{L}_{m}$, the whole training process is formulated as follows:
\begin{equation}
\label{eqn_5}
\hat{y_m} = FC_m(f_m)
\end{equation}
\begin{equation}
\label{eqn_6}
\mathcal{L}_{m} = |\hat{y_m} - y_m|
\end{equation}
\begin{equation}
\label{eqn_7}
\mathcal{L}_{reg} = \mathcal{L}_{f} + \sum_{m \in \{t,v,a\}}  \beta_m \, \mathcal{L}_{m} 
\end{equation}
where $FC_m$ is a fully connected layer and outputs unimodal sentiment score $\hat{y_m}$ and $y_m$ is the corresponding ground truth label. To modulate the contribution of each modality, we use $\beta_m$ to assign different weights to each modality. Therefore, the overall regression loss that needs to be optimized is $\mathcal{L}_{reg}$. 

\subsection{Unimodal Feature Representation}
Our model takes a video clip as input (including text, visual, and audio). For each modality sequence $m \in \{t,v,a\}$, we follow previous work~\cite{zadeh2017tensor,hazarika2020misa} to convert each modality sequence into a feature sequence $U_m$. We design unimodal feature extractors separately for different modalities to observe modality-specific features.

For text stream feature extraction, most works are based on pretrained language models. Experiments~\cite{sun2020learning,ismail2020improving,rahman2020integrating} show that different pretrained models exhibit different performance improvements. Early work uses GloVe~\cite{pennington2014glove} for feature embedding, and some work uses BERT~\cite{devlin2018bert} model. Following~\cite{hazarika2020misa,yu2021learning}, we choose a pretrained 12-layers BERT for text feature embedding, and the first word vector output by the last layer is used as the textual representation:
\begin{equation}
\label{eqn_8}
f_t = BERT(U_t;\theta_t^{bert}), \, f_t \in \mathbb{R}^{d_t^{\;'}}
\end{equation}
where $\theta_t^{bert}$ is the learnable parameter of BERT, $f_t$ is the text feature and $d_t^{\;'}$ is the embedded word vector dimension.

For vision and audio modalities, we use the toolkits~\cite{zadeh2017tensor,yu2021learning} for feature extraction, $U_v$ and $U_a$ represent the visual feature sequence and acoustic feature sequence, respectively. Two long short-term memory networks (LSTMs) are introduced to capture intramodality dynamics for each modality separately, and the final state hidden vector is represented as the entire sequence:
\begin{equation}
\label{eqn_9}
f_v = vLSTM(U_v;\theta_v^{lstm}), \, f_v \in \mathbb{R}^{d_v^{\;'}}
\end{equation}
\begin{equation}
\label{eqn_10}
f_a = aLSTM(U_a;\theta_a^{lstm}), \, f_a \in \mathbb{R}^{d_a^{\;'}}
\end{equation}
where $\theta_v^{lstm}$ and $\theta_a^{lstm}$ are learnable parameters, $f_v$ and $f_a$ are visual and audio features, respectively, $d_v^{\;'}$ and $d_a^{\;'}$ are the output dimensions of the two feature extractors.

To use the observed unimodal features for subsequent models, these unimodal features are projected into a low-dimensional feature space $\mathbb{R}^{d^*}$.
\begin{equation}
\label{eqn_11}
f_{m}^{*} = ReLU(W_m * f_m), \, m \in \{t,v,a\}
\end{equation}
where $f_{m}^*$ is the transformed feature, $W_m \in \mathbb{R}^{d^* \times d_m^{\;'}}$ is the transformation matrix, $ReLU(\cdot)$ is the ReLU activation function, and the bias term is omitted for simplicity.

We explore unimodal regression tasks to capture modality-specific features. Specifically, according to Eq.~(\ref{eqn_5}), three independent fully connected layers $FC_m$ are used to predict sentiment scores $\hat{y_m}$, and unimodal loss $\mathcal{L}_{m}$ is calculated with Eq.~(\ref{eqn_6}).

\subsection{Uncertainty Distribution Estimation}
Previous multimodal methods are based on deterministic embeddings, representing samples as deterministic points in the embedding space, thereby ignoring data uncertainty. We represent the feature embedding $\pmb{z}$ of sample $x$ in the latent space as a Gaussian distribution: $p(\pmb{z}|x)=\mathcal{N}(\pmb{\mu}, \pmb{\sigma}^2)$. Our uncertainty distribution estimation module directly takes the unimodal features $f_{m}^{*}$ as input and outputs the mean and variance:
\begin{equation}
\label{eqn_12}
\pmb{\mu}_m = MLP_{m}^{\mu}(f_{m}^{*}), \, \pmb{\sigma}_m = MLP_{m}^{\sigma}(f_{m}^{*})
\end{equation}
where $\pmb{\mu}_m$ represents the most likely feature embedding and $\pmb{\sigma}_m$ is interpreted as uncertainty, where larger variance values indicate higher uncertainty. Both $\pmb{\mu}_m$ and $\pmb{\sigma}_m$ are $D$-dimensional vectors representing the unimodal distribution of modality $m$. $MLP_{m}^{\mu}$ and $MLP_{m}^{\sigma}$ denote two multilayer perceptron (MLP) branches for predicting the mean and variance, respectively.

Now, we obtain unimodal representations $\mathcal{N}(\pmb{\mu}_m, \pmb{\sigma}_m^2)$ from which modality embeddings $\pmb{z}_m$ can be sampled. However, the sampling operation is not differentiable, preventing backpropagation of gradients during the training stage. Therefore, we consider using the reparameterization trick~\cite{kingma2013auto} to enable gradient propagation. Specifically, for each modality, random noise $\epsilon$ is sampled from a normal distribution $\mathcal{N}(\pmb{0},\pmb{I})$, and then $\pmb{z}_m$ is generated as the sampling result.
\begin{equation}
\label{eqn_13}
\pmb{z}_m = \pmb{\mu}_m + \epsilon \pmb{\sigma}_m, \, \epsilon \sim \mathcal{N}(\pmb{0},\pmb{I})
\end{equation}

To enable the estimated distribution to effectively capture modality-specific intrinsic information, we constrain it with a reconstruction loss to minimize the difference between unimodal features $f_{m}^{*}$ and modality embeddings $\pmb{z}_{m}$. First, MLP is used as a decoder to reconstruct: $\pmb{z}_{m}^{'} = MLP(\pmb{z}_{m}) \in \mathbb{R}^{d^*}$, and then the mean squared error (MSE) is taken as the reconstruction loss:
\begin{equation}
\label{eqn_14}
\mathcal{L}_{rec} = \frac{1}{3} \sum_{m \in \{t,v,a\}} \frac{||\pmb{z}_{m}^{'} - f_{m}^{*}||_{2}^{2}}{d^*} 
\end{equation}
where, $||\cdot||_{2}^{2}$ is the squared $L_2$-norm and $d^*$ is the number of elements in the feature.

Although the uncertainty distribution estimation module is capable of estimating the mean ($\pmb{\mu}_{m}$) and variance ($\pmb{\sigma}_{m}$) for modality $m$, too small a variance during training will cause our model to degenerate into a deterministic representation. To counteract this effect, the Kullback-Leibler (KL) divergence is introduced as a regularization term to constrain the estimated distribution ($\pmb{\mu}_{m}$ and $\pmb{\sigma}_{m}^2$) to approximate the normal distribution $\mathcal{N}(\pmb{0},\pmb{I})$. The value of KL represents the distance between two probability distributions.
\begin{align}
\label{eqn_15}
\mathcal{L}_{kl} &= \frac{1}{3} \sum_{m \in \{t,v,a\}} KL\left(\mathcal{N}(\pmb{\mu}_m,\pmb{\sigma}_m^2) \, || \, \mathcal{N}(\pmb{0},\pmb{I})\right) \nonumber \\
&=\frac{1}{3} \sum_{m \in \{t,v,a\}} \frac{1}{2} \bigl(\pmb{\mu}_{m^{2}} + \pmb{\sigma}_{m}^{2} - log\pmb{\sigma}_{m}^{2} - \pmb{1}\bigr)
\end{align}

Note that we aim to estimate unimodal uncertainty distributions under the cooperation of the two constraints. Guided by reconstruction loss $\mathcal{L}_{rec}$, our model is prompted to learn as much task-relevant information as possible. Guided by KL loss $\mathcal{L}_{kl}$, the model discourages lower variance. 

\subsection{Multimodal Distribution Fusion}
To date, unimodal representations for three modalities have been obtained: text ($\pmb{\mu}_t$, $\pmb{\sigma}_t^2$), visual ($\pmb{\mu}_v$, $\pmb{\sigma}_v^2$), and audio ($\pmb{\mu}_a$, $\pmb{\sigma}_a^2$). Conceptually, these distributions are treated as observations from different sensors and fused using Bayes' rule. Specifically, we illustrate with text modal distribution and visual modal distribution and estimate the fusion distribution parameters:
\begin{equation}
\label{eqn_16}
\pmb{\mu}_{f} = \frac{\pmb{\mu}_{t}\pmb{\sigma}_{v}^2 + \pmb{\mu}_{v}\pmb{\sigma}_{t}^2}{\pmb{\sigma}_{t}^2 + \pmb{\sigma}_{v}^2}, \, \pmb{\sigma}_{f}^2 = \frac{\pmb{\sigma}_{t}^2 \pmb{\sigma}_{v}^2}{\pmb{\sigma}_{t}^2 + \pmb{\sigma}_{v}^2}
\end{equation}
The fusion of the two distributions is a scaled Gaussian distribution, where $\pmb{\mu}_{f}$ can be seen as a weight of $\pmb{\mu}_{t}$ and $\pmb{\mu}_{v}$, and $\pmb{\sigma}_{f}^2$ is less than either $\pmb{\sigma}_{t}^2$ or $\pmb{\sigma}_{v}^2$. This result is in line with our expectation that different modality information is automatically weighed for each sample to obtain a more comprehensive representation with less uncertainty.

Unlike attention-based fusion methods~\cite{gu2018multimodal,liang2021attention,dong2022temporal}, our fusion module does not introduce additional learnable parameters. Notably, our fusion method is order-invariant, that is, it does not depend on the fusion order of modalities. Previous multimodal fusion methods focus on distinguishing primary and auxiliary modalities~\cite{wang2019words,sun2020learning,han2021improving}, with text as the primary modality and the remainder as auxiliary modalities. In the multimodal fusion stage, the text-audio modalities and the text-visual modalities are fused in stages, Then, the two fused results are integrated for final prediction. Instead of explicitly distinguishing these modalities, we weigh them indirectly through uncertainty estimation, which iteratively fuses unimodal distributions to obtain multimodal distributions $(\pmb{\mu}_{f},\pmb{\sigma}_{f}^2)$. Therefore, we can sample from the multimodal distribution to obtain fused multimodal features: $\pmb{z}_f = \pmb{\mu}_f + \varepsilon \pmb{\sigma}_f$, and $\varepsilon \sim \mathcal{N}(\pmb{0},\pmb{I})$.

To enrich multimodal information, we concatenate unimodal features ($f_t$, $f_v$, $f_a$) and multimodal features $\pmb{z}_f$ for sentiment prediction:
\begin{equation}
\label{eqn_17}
\hat{y}_f = FC_f([f_t;f_v;f_a;\pmb{z}_f])
\end{equation}
where $FC_f$ is a fully connected layer for sentiment regression and $[\cdot]$ is the concatenation operation. The multimodal loss $\mathcal{L}_{f}$ is obtained by Eq.~(\ref{eqn_3}). Therefore, the unimodal loss $\mathcal{L}_{m}$ and the multimodal loss $\mathcal{L}_{f}$ are combined to obtain the overall regression loss $\mathcal{L}_{reg}$.

\subsection{Ordinal Sentiment Regression}
\begin{figure}[t]
	\centering
	\includegraphics[width=8.8cm, height=3cm]{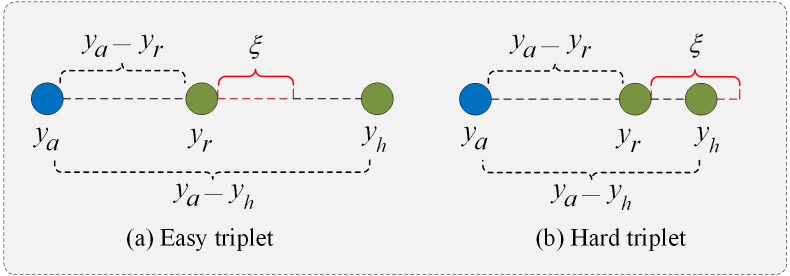}
	\caption{Triplet loss for regression. Blue circles represent anchors, and green circles represent reference points. (a) is the case where the samples are easily distinguishable, where the difference between $|y_a - y_{r}|$ and $|y_a - y_{h}|$ is much larger than $\xi$. (b) is the hard triplet, where the difference between $|y_a - y_{r}|$ and $|y_a - y_{h}|$ is smaller than $\xi$ and difficult to distinguish.}
	\label{triplet}
\end{figure}
Ordinal regression aims to learn a mapping function: $\mathcal{X} \rightarrow \mathcal{Y}$, where $\mathcal{X}$ is the data space and $\mathcal{Y} = \{\mathcal{C}_1, \mathcal{C}_2, \dots, \mathcal{C}_k\}$ is the target space containing $k$ categories. These label categories are in an ordering relationship: $\mathcal{C}_1 \prec \mathcal{C}_2 \prec \dots \prec \mathcal{C}_k$, where $\prec$ is an ordinal relation reflecting the relative position of the labels. We use ordinal regression to constrain the fused multimodal distributions $(\pmb{\mu}_{f},\pmb{\sigma}_{f}^2)$ so that the sentiment space has ordinal properties. Given a triplet distribution $\bigl(\mathcal{N}(\pmb{\mu}_{a}, \pmb{\sigma}_{a}^2), \, \mathcal{N}(\pmb{\mu}_{r}, \pmb{\sigma}_{r}^2), \, \mathcal{N}(\pmb{\mu}_{h}, \pmb{\sigma}_{h}^2)\bigr)$ and its associated label $\left(y_a, \, y_{r}, \, y_{h}\right)$. Assume a label ordinal relation: $\bigl|\mathcal{O}{(y_{a})} - \mathcal{O}{(y_{r})}\bigr| < \bigl|\mathcal{O}{(y_{a})} - \mathcal{O}{(y_{h})}\bigr|$, where $\mathcal{O}(\cdot)$ represents the position function of the label in the ordinal. Hence, the label relations are embedded in the multimodal distributions to preserve ordinal constraints:
\begin{align}
\label{eqn_18}
d\bigl(\mathcal{N}(\pmb{\mu}_{a}, \pmb{\sigma}_{a}^2), \mathcal{N}(\pmb{\mu}_{r}, \pmb{\sigma}_{r}^2)\bigr) &< d\bigl(\mathcal{N}(\pmb{\mu}_{a}, \pmb{\sigma}_{a}^2), \mathcal{N}(\pmb{\mu}_{h}, \pmb{\sigma}_{h}^2)\bigl) \nonumber \\ 
\Leftrightarrow \bigl|\mathcal{O}{(y_a)} - \mathcal{O}{(y_{r})}\bigr| &< \bigl|\mathcal{O}{(y_a)} - \mathcal{O}{(y_{h})}\bigr|
\end{align}
where $d(\cdot)$ is the distribution distance function. 

To accomplish the above constraints, a triplet loss is introduced. As illustrated in Fig.~\ref{triplet}, the triplet loss for regression does not need to construct positive and negative samples. Intuitively, samples with farther distances are easy to distinguish in Fig.~\ref{triplet}(a). We focus on constructing hard samples with similar sentiment scores in Fig.~\ref{triplet}(b). During the training phase, triplets are constructed for each sample in a mini-batch. Specifically, each sample is treated as anchor $a$, and one of the remaining samples except anchor $a$ is randomly selected as reference sample $r$. Subsequently, we find a perplexed sample as hard sample $h$ in the remaining samples, and the label relationship of these hard samples satisfies: $\bigl| |y_a - y_r| - |y_a - y_h| \bigr| < \xi$, where $\xi$ is a minimum value. Then, the triplet training set $\mathcal{B}$ is obtained. 

For simplicity, we use $\mathcal{P}_a$, $\mathcal{P}_r$ and $\mathcal{P}_h$ to simplify the distributions $\mathcal{N}(\pmb{\mu}_{a}, \pmb{\sigma}_{a}^2)$, $\mathcal{N}(\pmb{\mu}_{r}, \pmb{\sigma}_{r}^2)$, and $\mathcal{N}(\pmb{\mu}_{h}, \pmb{\sigma}_{h}^2)$, respectively.  ordinal loss is calculated as follows:
\begin{equation}
\label{eqn_19}
\mathcal{L}_{ord} = \frac{1}{|\mathcal{B}|} \sum_{ (a,r,h) \in \mathcal{B}} max\bigl(0, \, d(\mathcal{P}_a, \mathcal{P}_r) + \delta - d(\mathcal{P}_a, \mathcal{P}_h) \bigr)
\end{equation}
where $\delta$ is the margin and $|\mathcal{B}|$ is the number of samples in the training set. In a triplet, we aim to distinguish different sentiment categories. By introducing $\delta$, similar samples are drawn closer while different samples are pushed away. 

To select the distance function $d(\cdot)$, the Wasserstein distance is chosen to measure the distance of distributions. The Wasserstein distance is also known as the Earth Mover's distance (EMD), which defines the minimum cost required to change from one distribution to another. Taking the calculation of distributions $\mathcal{P}_a$ and $\mathcal{P}_r$ as an example, the distance $d\left(\mathcal{P}_a, \mathcal{P}_r\right)$ is calculated:
\begin{align}
\label{eqn_20}
d\left(\mathcal{P}_a, \mathcal{P}_r\right) :&= W_2\left(\mathcal{N}(\pmb{\mu}_{a}, \pmb{\sigma}_{a}^2), \, \mathcal{N}(\pmb{\mu}_{r}, \pmb{\sigma}_{r}^2) \right)^2 \nonumber \\ 
&= \sum_{j=1}^D{\left[(\pmb{\mu}_{a,j} - \pmb{\mu}_{r,j})^2 + (\pmb{\sigma}_{a,j} - \pmb{\sigma}_{r,j})^2\right]}
\end{align}
where $W_2(\cdot)$ represents the 2nd-order Wasserstein distance, $D$ represents the number of feature channels, $\pmb{\mu}_{a,j}$ and $\pmb{\sigma}_{r,j}$ represent the $j$-th dimension of $\pmb{\mu}_{a}$ and $\pmb{\sigma}_{r}$, respectively.

\subsection{Training }
We combine regression loss $\mathcal{L}_{reg}$ (multimodal and unimodal loss), reconstruction loss $\mathcal{L}_{rec}$, KL loss $\mathcal{L}_{kl}$, and ordinal loss $\mathcal{L}_{ord}$ into the model. The overall loss is:
\begin{equation}
	\label{eqn}
	\mathcal{L} = \mathcal{L}_{reg} + \lambda_1\mathcal{L}_{rec} + \lambda_2\mathcal{L}_{kl} + \lambda_3\mathcal{L}_{ord}
\end{equation}
where $\lambda_1$, $\lambda_2$ and $\lambda_3$ are hyperparameters used to trade off different loss components.

\section{Experiments}
In this section, experiments are conducted to evaluate the performance of TMSON for sentiment analysis. The performance of TMSON is compared on three benchmark datasets, CMU-MOSI~\cite{zadeh2016mosi}, CMU-MOSEI~\cite{zadeh2018multimodal} and SIMS~\cite{yu2020ch}. We further explore each module to demonstrate the effectiveness of TMSON. In addition, we analyze the robustness of TMSON to demonstrate its superiority. Finally, we conduct correlation visualization experiments to demonstrate the effectiveness of ordinal regression and the interpretability of multimodal fusion.

\subsection{Datasets}
\subsubsection{CMU-MOSI}
A video opinion dataset from YouTube movie reviews, which is used as a benchmark dataset for multimodal sentiment analysis. CMU-MOSI is collected from 93 videos, including 2199 short video clips, each with an average length of 4.2 seconds. Video sentiment annotation follows the Stanford Sentiment Treebank~\cite{socher2013recursive}, where sentiment is annotated according to a seven-step Likert scale. The intensity of these sentiments is in the range [-3: highly negative, -2: negative, -1: weakly negative, 0: neutral, +1: weakly positive, +2: positive, +3: highly positive]. Of the entire dataset, 1284 utterances are used as the training set, 229 utterances are used as the validation set, and 686 utterances are used as the test set. 

\subsubsection{CMU-MOSEI}
This dataset is an extension of CMU-MOSI and is the largest multimodal sentiment analysis dataset with 3228 videos from 1000 speakers on 250 topics. CMU-MOSEI closely follows the sentiment annotation method of CMU-MOSI, and the sentiment annotation of each utterance is in the range [-3, +3], where -3 is the strongest negative emotion and +3 is the strongest positive emotion. There are 23453 utterances after processing; 16265 utterances are used as the training set, 2545 fragments are used as the validation set, and 4643 utterances are used as the test set.

\subsubsection{SIMS}
A Chinese multimodal sentiment dataset with fine-grained annotations on both unimodal and multimodal. SIMS is collected from movies, TV serials, and variety shows. It contains 60 raw videos and is parsed into 2281 video clips, with an average of 15 words per clip. The sentiment annotation is \{-1.0, -0.8, -0.6, -0.4, -0.2, 0.0, 0.2, 0.4, 0.6, 0.8, 1.0\}. Furthermore, these values are divided into 5 categories: negative \{-1.0, -0.8\}, weakly negative \{-0.6, -0.4, -0.2\}, neutral \{0.0\}, weakly positive \{0.2, 0.4, 0.6\}, and positive \{0.8, 1.0\}. We use 1368 utterances as the training set, 456 utterances as the validation set, and 457 utterances as the test set. 

\subsection{Baselines}
In this section, we conduct a fair comparison with the following baselines and state-of-the-art multimodal sentiment analysis models.

\subsubsection{EF\_LSTM and LF\_DNN}
Early fusion LSTM (EF-LSTM)~\cite{socher2013recursive}, which concatenates multiple modalities at the input level, after which the concatenated features are fed to an LSTM for final prediction. Conversely, late fusion LSTM (LF-LSTM)~\cite{socher2013recursive} uses independent LSTMs to extract unimodal features and then combines unimodal decisions with a voting mechanism.

\subsubsection{TFN and LMF}
Tensor fusion network (TFN)~\cite{zadeh2017tensor} is the first to formulate the multimodal sentiment analysis problem as modeling intra- and inter-modal dynamics. TFN uses independent LSTMs to capture each modal information separately and then applies the threefold Cartesian product from the modal embeddings to model intermodal interactions. Low-rank multimodal fusion (LMF)~\cite{liu2018efficient} decomposes the weights into low-rank factors, thereby reducing the number of parameters in the model, and the computational complexity scales linearly with the number of modalities.

\subsubsection{MFM}
Multimodal factorization model (MFM)~\cite{tsai2019learning} introduces a factorization model that decomposes the multimodal representation into two sets of independent factors: the multimodal discriminant factor and the modality-specific generative factor. MFM is able to learn multimodal representations and reconstruct missing modalities.

\subsubsection{RAVEN}
Recurrent attended variation embedding network (RAVEN)~\cite{wang2019words} considers that human expressions are accompanied by verbal and nonverbal behaviors and thus that literal language needs to be comprehensively considered in conjunction with external patterns. RAVEN changes the representation of words by modeling nonlinguistic representations.

\subsubsection{MCTN}
Multimodal cyclic translation model (MCTN)~\cite{pham2019found}, which adopts the Seq2Seq model, introduces a cycle consistency loss between modality transitions to ensure that the joint representation retains the maximum information from all modalities and learns a robust joint representation.

\subsubsection{ICCN}
Interaction canonical correlation network (ICCN)~\cite{sun2020learning}, which takes text modality as the primary modality and learns text-based audio features and text-based video features. Finally, the learned bimodal representation is input into a canonical correlation analysis (CCA) network to model the interaction of the three modalities.

\subsubsection{CIA}
Context-aware interactive attention (CIA)~\cite{chauhan2019context} is a sentiment analysis model based on recurrent neural networks that uses autoencoders to capture intermodality interactions. Furthermore, CIA introduces a context-aware attention module to explore the relationship between adjacent utterances.

\subsubsection{MAG-BERT}
Multimodal adaptation Gate BERT (MAG-BERT)~\cite{rahman2020integrating} proposes an attachment that combines with language pretrained models, such as BERT~\cite{devlin2018bert} and XLNet~\cite{yang2019xlnet}, to enable pretrained models to accept multimodal inputs during fine-tuning.

\subsubsection{Mu-Net}
Multilogue-net~\cite{shenoy2020multilogue} proposes an end-to-end RNN architecture that exploits and captures dependencies between dialog context, listener and speaker emotional states across all modalities and learns relevance and relationships between available modalities.

\subsubsection{GraphCAGE}
Graph Capsule Aggregation (GraphCAGE)~\cite{wu2021graph} uses graph networks and capsule networks to model unaligned multimodal sequences. GraphCAGE converts sequence data into graphs, avoiding the problem of vanishing or exploding gradients.

\subsubsection{Mult}
Multimodal Transformer (MulT)~\cite{tsai2019multimodal} focuses on the long-term dependencies of unaligned multimodal sequences. MulT introduces a cross-modal Transformer to transform one modality to another, focusing on interactions between multimodal sequences across different time steps.

\subsubsection{MISA}
MISA~\cite{hazarika2020misa}, which projects each modality into two subspaces, learns modality-invariant and modality-specific representations, after which these representations are fused for multimodal prediction. 

\subsubsection{Self-MM}
Self-MM~\cite{yu2021learning} proposes a self-supervised label generation module to obtain unimodal labels. Subsequently, the multimodal and unimodal tasks are jointly trained to learn consistency and difference.

\subsubsection{MMIM}
MultiModal InfoMax (MMIM)~\cite{han2021improving} maximizes the mutual information in unimodal input pairs and between multimodal fusion results and unimodal input to preserve task-relevant information. 

\subsection{Experimental Details}
\begin{table}[t]
	\renewcommand{\arraystretch}{1.2}
	\caption{The best performing hyperparameters on different datasets.}
	\label{hyperparameters}
	\centering
	\resizebox{8.5cm}{3.4cm}{
		\begin{tabular}{c c c c}
			\toprule
			Item & CMU-MOSI & CUM-MOSEI & SIMS \\
			\midrule
			batch size & 16 & 32 & 32 \\
			feature dim ($d_t,d_v,d_a$) & (768,20,5) & (768,35,74) & (768,709,33) \\
			$BERT$ dropout & 0.1 & 0.1 & 0.1\\
			$BERT$ weight decay & 1e-3 & 1e-3 & 1e-3\\
			$v\_LSTM$ hidden dim $d_v^{\;'}$ & 32 & 32 & 64 \\
			$v\_LSTM$ weight decay & 1e-3 & 0.0 & 1e-2\\
			$a\_LSTM$ hidden dim $d_a^{\;'}$ & 16 & 16 & 16 \\
			$a\_LSTM$ weight decay & 1e-3 & 0.0 & 1e-2\\
			other weight decay & 1e-3 & 1e-2 & 1e-3\\
			$d^*$ & 128 & 128 & 128\\
			distribution dim $D$ & 64 & 64 & 64\\
			$FC_f$ hidden dim & 128 & 128 & 128\\
			$FC_t$ hidden dim & (32 $\rightarrow$ 1) & (32 $\rightarrow$ 1) & (64 $\rightarrow$ 1) \\
			$FC_v$ hidden dim & (32 $\rightarrow$ 1) & (32 $\rightarrow$ 1) & (32 $\rightarrow$ 1) \\
			$FC_a$ hidden dim & (16 $\rightarrow$ 1) & (16 $\rightarrow$ 1) & (16 $\rightarrow$ 1) \\
			$FC_t$ dropout & 0.1 & 0.0 & 0.1 \\
			$FC_a$ dropout & 0.1 & 0.0 & 0.1 \\
			$FC_f$ dropout & 0.0 & 0.1 & 0.0 \\
			\bottomrule
	\end{tabular}}
\end{table}

\para{Framework implementation}. For fair comparison, we use an open-source platform MMSA\footnote{https://github.com/thuiar/MMSA} that integrates existing multimodal sentiment analysis methods. All experiments are conducted on this platform. TMSON is developed with CUDA 11.6 and PyTorch 1.12 on a single RTX 3090.

\para{Feature extraction}. We follow the feature extraction methods such as~\cite{yu2021learning,han2021improving}. Specifically,  CMU-MultimodalSDK\footnote{https://github.com/A2Zadeh/CMU-MultimodalSDK}, Facet\footnote{iMotions 2017. https://imotions.com/} and COVAREP~\cite{2014COVAREP} are used to extract text, visual, and audio features, respectively. To enable a fair comparison, the three benchmark datasets are identical to those used in MMSA. For CMU-MOSI, the input feature dimensions for text, visual, and audio modalities are 768, 20, and 5, respectively. For CMU-MOSEI, the input dimensions for text, visual, and audio are 768, 35, and 74, respectively. For SIMS, the input dimensions for text, visual, and audio  are 768, 709, and 33, respectively.

\para{Evaluation metrics}. To maintain consistency with previous work, the same evaluation metrics are adopted~\cite{zadeh2017tensor}. The evaluation is carried out from two main aspects: regression and classification. For regression, we use the mean absolute error (MAE) to measure the difference between the prediction and ground truth and Pearson correlation (Corr) to measure the degree of prediction skew. For regression, CMU-MOSI and CMU-MOSEI are evaluated with the same metrics, namely, binary classification accuracy (Acc-2) for positive and negative prediction evaluation, F1 score for positive/negative and negative/non-negative classification evaluation, and 7-class classification accuracy (Acc-7) predicts the proportion of 7 intervals that correctly fall within [-3, +3]. For SIMS, Acc-2, Acc-3, Acc-5, MAE, F1 and Corr are used as evaluation metrics. Among them, Acc-2 is used to evaluate two class intervals: [-1, 0] and [0, 1]; Acc-3 is used to evaluate three class intervals: [-1, -0.1], [-0.1, 0.1] and [0.1, 1.0]; and Acc-5 evaluates five class intervals: [-1, -0.7], [-0.7, -0.1], [-0.1, 0.1], [0.1, 0.7] and [0.7, 1].

\para{Parameter details}. TMSON uses Adam as the optimizer with an initial learning rate of 5e-5 for BERT, 5e-3 for visual and audio feature extractors, and 1e-3 for other parameters. For unimodal loss weight $\beta_m$,  $\beta_t=0.8$, $\beta_v=0.1$ and $\beta_a=0.1$. The hyperparameters $\lambda_1$, $\lambda_2$, and $\lambda_3$ are set to 0.1, 0.01, and 0.5, respectively. The margin $\delta$ for ordinal regression is set to 1. For the multitask setting, we set the unimodal label to be the same as the multimodal label. More parameter configurations on different datasets are shown in Table~\ref{hyperparameters}.

\subsection{Experimental Results}
\begin{table*}[t]
	\renewcommand{\arraystretch}{1.2}
	\caption{Comparison with baselines on the CMU-MOSI and CMU-MOSEI datasets. Note that the left side of ``/" is ``negative/non-negative" and the right is ``negative/positive". The best results are highlighted in bold.}
	\label{sentiment_result}
	\centering
	\begin{tabular}{c c c c c c c c c c c c}
		\toprule
		& \multicolumn{5}{c}{CMU-MOSI} & & \multicolumn{5}{c}{CMU-MOSEI}\\ 
		\cmidrule{2-6}\cmidrule{8-12}
		Models & Acc-2 & F1 & Acc-7 & MAE & Corr & & Acc-2 & F1 & Acc-7 & MAE & Corr\\
		\midrule
		TFN~\cite{zadeh2017tensor} & -/80.8 & -/80.7 & 34.9 & 0.901 & 0.698 & & -/82.5 & -/82.1 & 50.2 & 0.593 & 0.700\\
		LMF~\cite{liu2018efficient} & -/82.5 & -/82.4 & 33.2 & 0.917 & 0.695 & & -/82.0 & -/82.1 & 48.0 & 0.623 & 0.677\\
		MFM~\cite{tsai2019learning} & -/81.7 & -/81.6 & 35.4 & 0.877 & 0.706 & & -/84.4 & -/84.3 & 51.3 & 0.568 & 0.717\\
		ICCN~\cite{pham2019found} & -/83.0 & -/83.0 & 39.0 & 0.862 & 0.714 & & -/84.2 & -/84.2 & 51.6 & 0.565 & 0.713\\
		RAVEN~\cite{wang2019words} & 76.6/- & 78.0/- & 33.2 & 0.915 & 0.691 & & 79.5/- & 79.1/- & 50.0 & 0.614 & 0.662\\
		MCTN~\cite{pham2019found} & 79.3/- & 79.1/- &35.6 & 0.909 & 0.676 & & 79.8/- &80.6/- & 49.6 & 0.609 & 0.670\\
		CIA~\cite{chauhan2019context} &79.5/- &79.8/- & 38.9 & 0.914 & 0.689 & & 78.2/- & 80.4/- & 50.1 & 0.680 & 0.662\\
		GraphCAGE~\cite{wu2021graph} &-/82.1 &-82.1 & 35.4 & 0.933 & 0.684 & & -/81.7 & -/81.8 & 48.9 & 0.609 & 0.670\\
		Mu-net~\cite{shenoy2020multilogue} & -/80.1 & -/81.2 & - & - & - & & -/80.0 & -/82.1 & - & 0.590 & 0.50\\
		MulT~\cite{tsai2019multimodal} & -/84.1 & 80.6/83.9 & - & 0.861 & 0.711 & & -/82.5 & -/82.3 & - & 0.580 & 0.703\\
		MISA~\cite{hazarika2020misa} & 81.8/83.4 & 81.7/83.6 & 42.3 & 0.783 & 0.761 & & 83.6/85.5 & 83.8/85.3 & 52.2 & 0.555 & 0.756\\
		MAG-BERT~\cite{rahman2020integrating} & 82.5/84.3 & 82.6/84.3 & - & 0.731 & 0.789 & & 83.8/85.2 & 83.7/85.1 & - & 0.539 & 0.753\\
		Self-MM~\cite{yu2021learning} & 84.0/86.0 & 84.4/86.0 & - & 0.713 & 0.798 & & 83.8/85.2 & 83.7/85.1 & - & 0.539 & 0.753\\
		MMIM~\cite{han2021improving} & 84.1/86.1 & 84.0/86.0 & 46.7 & 0.700 & 0.800 & & 82.2/86.0 & 82.7/85.9 & 54.2 &\textbf{0.526} & \textbf{0.772}\\
		\midrule
		TMSON & \textbf{85.4}/\textbf{87.2} & \textbf{85.4}/\textbf{87.2} & \textbf{47.4} & \textbf{0.687} & \textbf{0.809} & & \textbf{85.2}/\textbf{86.4} & \textbf{85.3}/\textbf{86.2} & \textbf{55.6} & \textbf{0.526} & 0.766\\
		\bottomrule
	\end{tabular}
\end{table*}

\begin{table}[t]
	\renewcommand{\arraystretch}{1.2}
	\caption{Comparison with baselines on the SIMS dataset. MLF\_DNN and MTFN are developed by MMSA.}
	\label{SIMS}
	\centering
	\begin{tabular}{c c c c c c c }
		\toprule
		Models & Acc-2 & Acc-3 & Acc-5 & F1 & MAE & Corr \\
		\midrule
		EF\_LSTM~\cite{socher2013recursive} & 69.4 & 54.3 & 21.2 & 56.8 & 0.591 & 0.055\\
		LF\_DNN~\cite{socher2013recursive} & 77.2 & 64.3 & 39.7 & 77.3 & 0.446 & 0.555\\
		TFN~\cite{zadeh2017tensor} & 78.4 & 65.1 & 39.3 & 78.6 & 0.432 & 0.591\\
		LMF~\cite{liu2018efficient} & 77.8 & 64.7 & 40.5 & 77.9 & 0.411 & 0.576\\
		MFM~\cite{tsai2019learning} & 77.9 & 65.7 & 39.5 & 77.9 & 0.435 & 0.582\\
		Graph-MFN~\cite{zadeh2018multi} & 78.8 & 65.7 & 39.8 & 78.2 &0.445 & 0.578\\
		MulT~\cite{tsai2019multimodal} & 78.6 & 64.8 & 37.9 & 79.7 & 0.453 & 0.564\\
		MLF\_DNN & 80.4 & \textbf{69.4} & 40.2 & 80.3 & 0.396 & 0.665\\
		MTFN & 81.1 & 68.8 & 40.3 & 81.0 & \textbf{0.395} & \textbf{0.666}\\
		Self-MM~\cite{yu2021learning} & 80.0 & 65.5 & 41.5 & 80.4 & 0.425 & 0.592\\
		\midrule
		TMSON & \textbf{81.4} & 66.4 & \textbf{45.3} & \textbf{81.2} & 0.417 & 0.658\\
		\bottomrule
	\end{tabular}
\end{table}

The proposed TMSON is compared with baseline models. The experimental results demonstrate that TMSON outperforms on multiple metrics. In multimodal sentiment analysis, text modality is often regarded as the primary modality, so the feature representation of text modality is crucial. Note that in Table~\ref{sentiment_result}, most of the baseline models (such as TFN~\cite{zadeh2017tensor}, LMF~\cite{liu2018efficient}, MulT~\cite{tsai2019multimodal}, Self-MM~\cite{yu2021learning}, etc.) adopt BERT as the text encoder. In the experimental setting, a fair comparison is achieved using the off-the-shelf pretrained model BERT. 

We conduct comprehensive experiments on the CMU-MOSI, CMU-MOSEI and SIMS datasets, and TMSON outperforms these competitors in most cases. More specifically, as shown in Table~\ref{sentiment_result}, TMSON outperforms other baseline models on CMU-MOSI. Especially in the two metrics Acc-2 and F1 scores, TMSON significantly improves 1.3\%/1.1\% and 1.4\%/1.2\%, respectively, and shows competitive performance on other metrics. The two current state-of-the-art models are Self-MM and MMIM. The former benefits from unimodal fine-grained annotation generation and multitask learning, and the latter benefits from learning better multimodal representations with hierarchical mutual information constraints. Our TMSON introduces uncertainty estimates and ordinal constraints, where uncertainty estimates can automatically trade off the quality of different modalities, and ordinal constraints force the model to establish the ordinal relationship of emotions.

By using the CMU-MOSEI dataset, as shown in Table~\ref{sentiment_result}, it is confirmed that TMSON has the most significant improvement in (negative) Acc-2, (negative) F1 score and Acc-7, with increases of 3\%, 2.6\%, and 1.4\%, respectively. For other metrics, TMSON achieves close performance. With more training samples, TMSON adequately captures the sentiment ordinal relationships.More information on the TMSON ablation experiments is provided in Section~\ref{ablation_study}.

For the SIMS dataset, on which most baseline models do not provide performance, we compare the performance reproduced by MMSA\footnote{https://github.com/thuiar/MMSA/blob/master/results/result-stat.md}. The results are shown in Table~\ref{SIMS}. TMSON achieves comparable performance. Note that MLF\_DNN and MTFN are both multitask models using fine-grained unimodal manual annotations and multimodal annotations, whereas TMSON only uses multimodal annotations.

\subsection{Ablation Study}\label{ablation_study}
\subsubsection{Validation for Different Modalities}
\label{different_modalities}
To further explore TMSON, we explore the performance under different modality combinations. The experimental results are shown in Table~\ref{table4}. In unimodal experiments, ``T", ``V", and ``A" were used to represent only text, visual, and audio as input. Practical results confirm that the text modality is the most predictive, while the visual modality has the worst performance. The reason is that text modalities benefit from pretrained models and are able to obtain more efficient feature representations. However, visual and audio modalities are limited by artificial features, and the parameters of the feature extractor are randomly initialized, so more efficient feature representations need to be explored.

Then, bimodal experiments are conducted to analyze the influence between different modalities. Experimental results show the improvement of sentiment prediction with different modality combinations. Bimodal input shows better performance than unimodal input. Comparing ``V+T" and ``A+T" shows that the audio modality improves the most over the text modality. We speculate that the reason for this improvement may be that language and audio usually occur simultaneously when humans express opinions, and sound can provide more complementary information than facial expressions.

Finally, trimodal experiments ``V+A+T" are conducted, and the performance improvement of the trimodal is the most significant compared with unimodal input and bimodal input. This result also confirms that by learning complementary information between multiple modalities, it is possible to promote the consistent representation of multiple modalities.

\begin{table}[t]
	\renewcommand{\arraystretch}{1.2}
	\caption{Comparison of different modalities on CMU-MOSI.}
	\label{table4}
	\centering
	\begin{tabular}{c c c c c c }
		\toprule
		modality & Acc-2 & F1 & Acc-7 & MAE & Corr \\
		\midrule
		V & 53.9/53.2 & 53.6/53.0 & 16.9 & 1.437 & 0.122\\
		A & 62.5/63.9 & 61.5/63.1 & 17.1 & 1.361 & 0.278\\
		T & 82.4/83.9 & 82.3/84.1 & 45.8 & 0.704 & 0.803\\
		\midrule
		V+A & 67.2/67.1 & 65.7/65.5 & 26.5 & 1.125 & 0.433\\
		V+T & 82.9/84.8 & 82.9/84.9 & 46.7 & 0.701 & 0.800\\
		A+T & 83.8/86.0 & 83.5/85.9 & 46.4 & 0.709 & 0.801\\
		\midrule
		V+A+T & \textbf{85.4}/\textbf{87.2} & \textbf{85.4}/\textbf{87.2} & \textbf{47.4} & \textbf{0.687} & \textbf{0.809}\\
		\bottomrule
	\end{tabular}
\end{table}

\subsubsection{Discussion of Fusion Strategies}
Uncertain fusion methods on the CMU-MOSI dataset are compared, as shown in Table~\ref{fusin_methods}, specifically involving Sum, Average, PFE, and Bayesian fusion. Experimental results show that Bayesian fusion achieves the best performance. Among them, the sum operation directly adds each modality, resulting in the accumulation of deviations. The average operation takes the mean value in each dimension, which is moderate. PFE takes a dimensionwise minimum to obtain the variance and then integrates means according to the quality of the variance. Therefore, PFE focuses on the modality with the least uncertainty, which is one-sided. Bayesian fusion takes into account the variance of each modality and obtains a more comprehensive distribution fusion. 
\begin{table}[t]
	\renewcommand{\arraystretch}{1.3}
	\caption{Uncertainty fusion strategies.}
	\label{fusin_methods}
	\centering
	\begin{tabular}{c c c c c c }
		\toprule
		method & Acc-2 & F1 & Acc-7 & MAE & Corr \\
		\midrule
		Sum & 82.4/84.5 & 82.2/84.4 & \textbf{47.6} & 0.704 & 0.796\\
		Average & 82.8/85.4 & 82.6/85.3 & 46.9 & 0.702 & 0.796\\
		PFE~\cite{shi2019probabilistic} & 83.8/86.1 & 83.7/85.9 &46.9 &0.706& 0.801\\
		Bayesian fusion & \textbf{85.4}/\textbf{87.2} & \textbf{85.4}/\textbf{87.2} & 47.4 & \textbf{0.687} & \textbf{0.809}\\
		\bottomrule
	\end{tabular}
\end{table}

\subsubsection{Ablation Studies of Functional Modules}
\begin{table}[t]
	\renewcommand{\arraystretch}{1.2}
	\caption{Ablation studies on the CMU-MOSI dataset.}
	\label{functional_module}
	\centering
	\begin{tabular}{c c c c c }
		\toprule
		& F1 & Acc-7 & MAE & Corr \\
		\midrule
		$\mathcal{L}_{f}$ & 82.40/84.25 & 45.69 & 0.718 & 0.762\\
		
		$\mathcal{L}_{reg}$ & 82.75/84.61 & 46.58 & 0.706 & 0.772\\
		
		$\mathcal{L}_{reg}$ + $\mathcal{L}_{rec}$ & 82.96/85.15 & \textbf{47.52} & 0.699 & 0.791\\
		
		$\mathcal{L}_{reg}$ + $\mathcal{L}_{ord}$ & 83.09/85.49 &46.79 &0.701 & 0.803\\
		
		$\mathcal{L}_{reg}$ + $\mathcal{L}_{rec}$ + $\mathcal{L}_{kl}$ & 83.59/85.89 & 46.36 & 0.689 & 0.796\\
		
		$\mathcal{L}_{reg}$ + $\mathcal{L}_{rec}$ + $\mathcal{L}_{ord}$ & 84.28/86.38 & 46.63 & 0.697 & 0.802\\
		
		$\mathcal{L}_{reg}$ +  $\mathcal{L}_{rec}$ + $\mathcal{L}_{ord}$ + $\mathcal{L}_{kl}$ & \textbf{85.37}/\textbf{87.19} & 47.43 & \textbf{0.687} & \textbf{0.809}\\
		\bottomrule
	\end{tabular}
\end{table}

Typical multimodal sentiment analysis models use regression loss $\mathcal{L}_{reg}$ as a guide function. In TMSON, reconstruction loss $\mathcal{L}_{rec}$, KL loss $\mathcal{L}_{kl}$ and ordinal loss $\mathcal{L}_{ord}$ are introduced. Ablation experiments are performed on these functional modules to verify their performance. Among them, regression loss is essential, which guides the entire model for sentiment regression. Thus, we use regression loss as the benchmark for comparing each functional module. 

Table~\ref{functional_module} shows the performance of each module on the CMU-MOSI dataset, evaluated with Acc-7, F1 score, MAE and Corr metrics. We first investigate multitask learning and decompose regression loss into unimodal loss ($\mathcal{L}_{m}$) and multimodal loss ($\mathcal{L}_{f}$). Comparing the two experiments of $\mathcal{L}_{f}$ and $\mathcal{L}_{reg}$ shows that the performance improves after combining with unimodal loss. Two groups of experiments involving ``$\mathcal{L}_{reg}$ + $\mathcal{L}_{rec}$" and ``$\mathcal{L}_{reg}$ + $\mathcal{L}_{ord}$" show that adding $\mathcal{L}_{rec}$ or $\mathcal{L}_{ord}$ improves performance. Furthermore, comparing ``$\mathcal{L}_{reg}$ + $\mathcal{L}_{rec}$" and ``$\mathcal{L}_{reg}$ + $\mathcal{L}_{rec}$ + $\mathcal{L}_{kl}$", KL loss further improves performance and prevents the distribution representation from degenerating into a deterministic representation, which is in line with our expectation. Compared with ``$\mathcal{L}_{reg}$ + $\mathcal{L}_{rec}$", ``$\mathcal{L}_{reg}$ + $\mathcal{L}_{rec}$ + $\mathcal{L}_{ord}$" integrates ordinal loss, which significantly improves the performance in F1 score. Finally, all modules are integrated and achieve the best performance. 

\subsubsection{Weight Analysis of Loss Terms}
\begin{figure}[t]
	\centering
	\includegraphics[width=9.cm, height=4cm]{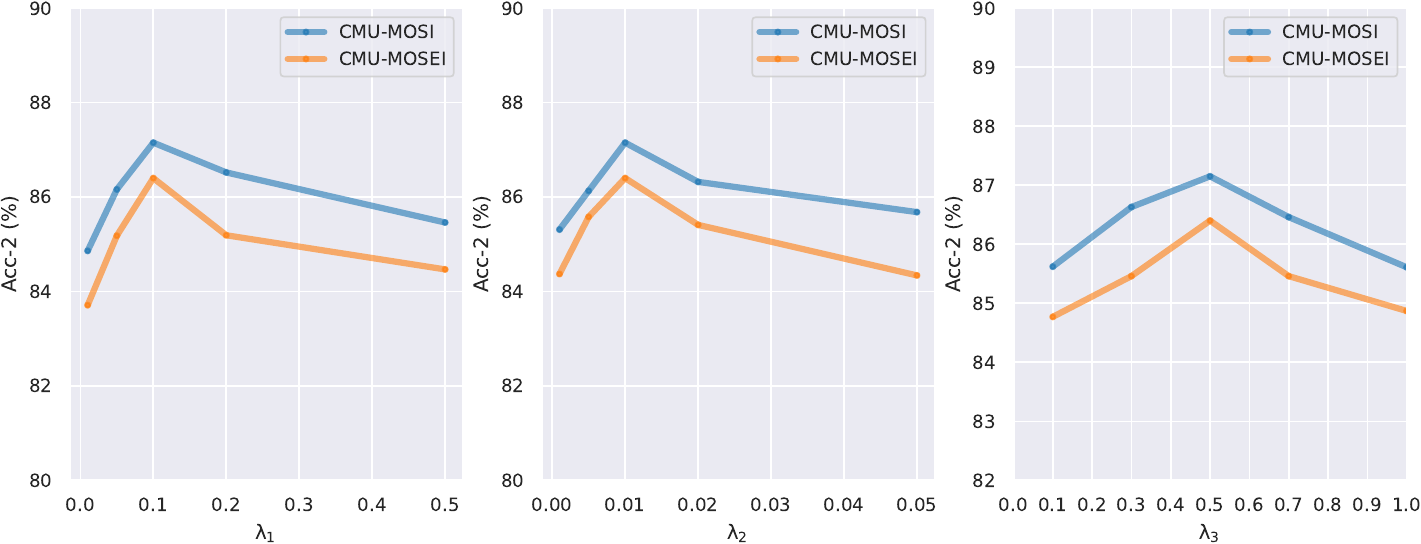}
	\caption{Weight analysis of different loss terms.}
	\label{loss_weight}
\end{figure}
TMSON consists of multiple loss terms, and we discuss the weights of each loss term to understand their contributions. Fig.~\ref{loss_weight} shows the performance of the three loss terms on CMU-MOSI and CMU-MOSEI. We select a moderate weight range of [0, 0.5] for $\lambda_1$ to emphasize sample reconstruction and a smaller weight range of [0, 0.05] to constrain the learned distribution for $\lambda_2$. $\lambda_1$ and $\lambda_2$ exhibit similar performance, and weights that are too small or too large lead to performance degradation. Therefore, moderate weights achieve the best performance. For $\lambda_3$, considering the importance of the ordinal relationship, we choose a larger weight range of [0, 1]. The best performance is achieved when the weight is 0.5, and increasing the weight further leads to performance degradation.

\subsubsection{Capturing Uncertainty}
\label{estimate_uncertainty}
\begin{figure*}[t]
	\centering
	\subfloat[fusion distribution.]{\includegraphics[width=4.5cm, height=4cm]{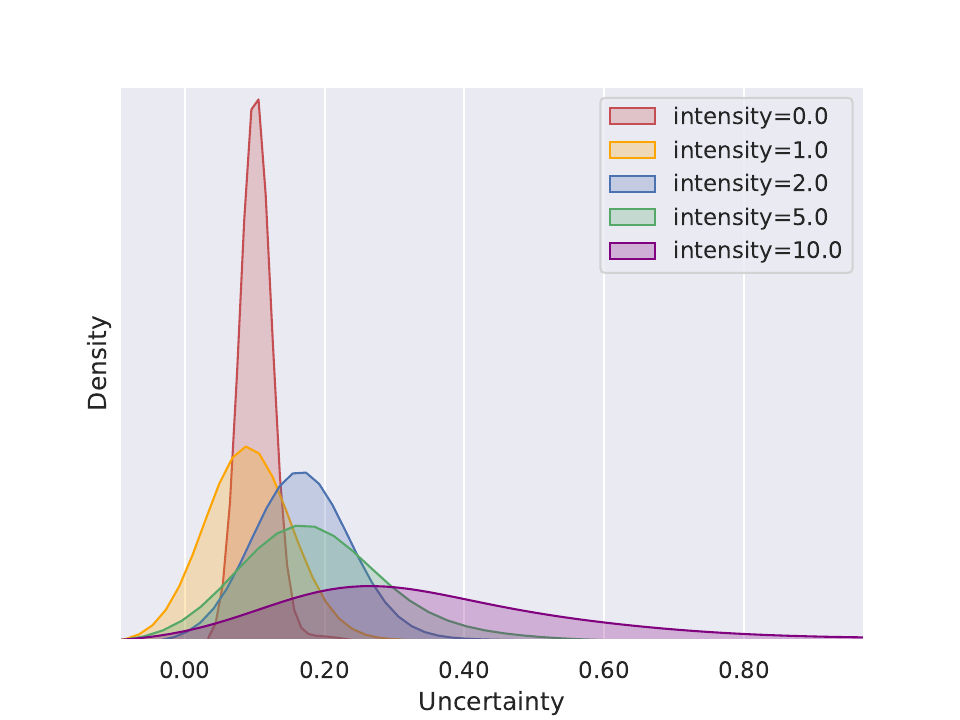}%
		\label{fig_f}}
	\hfil
	\subfloat[text distribution.]{\includegraphics[width=4.5cm, height=4cm]{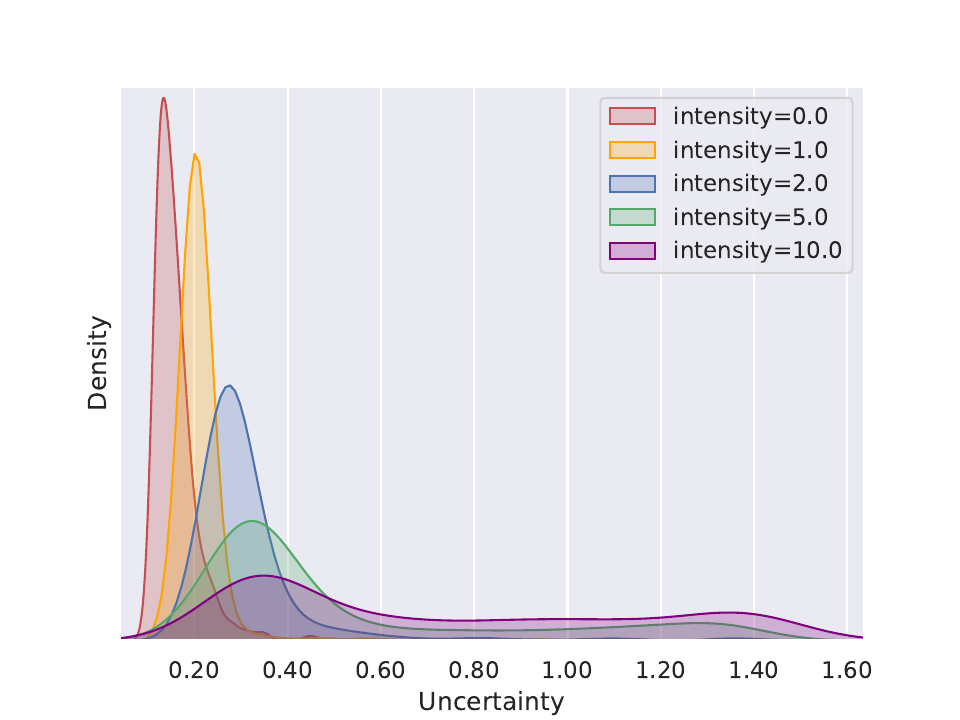}%
		\label{fig_t}}
	\hfil
	\subfloat[visual distribution.]{\includegraphics[width=4.5cm, height=4cm]{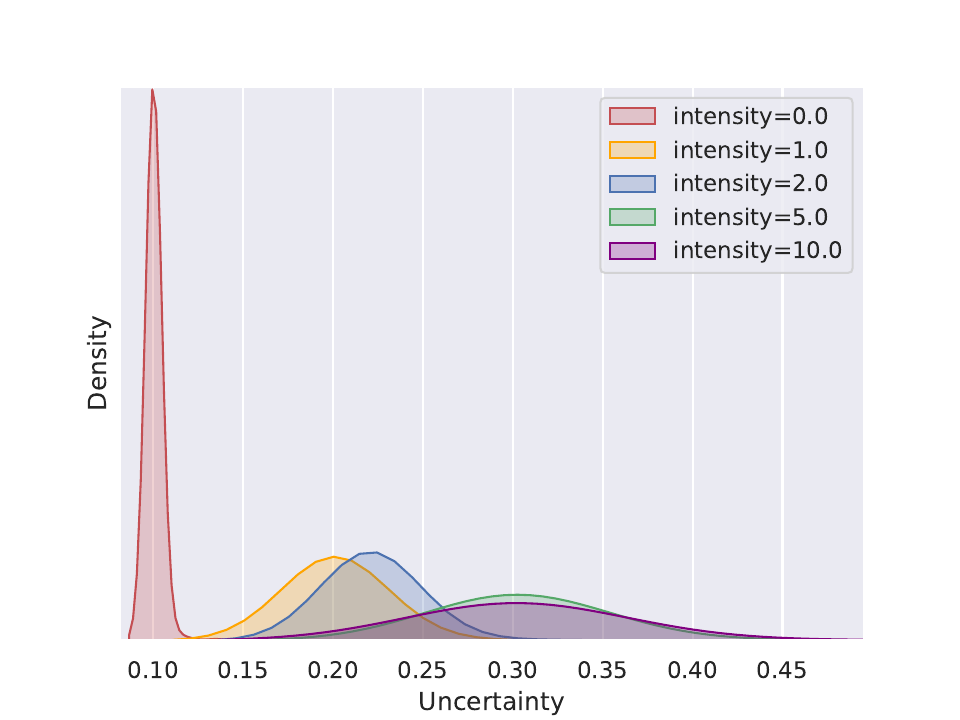}%
		\label{fig_v}}
	\hfil
	\subfloat[audio distribution.]{\includegraphics[width=4.5cm, height=4cm]{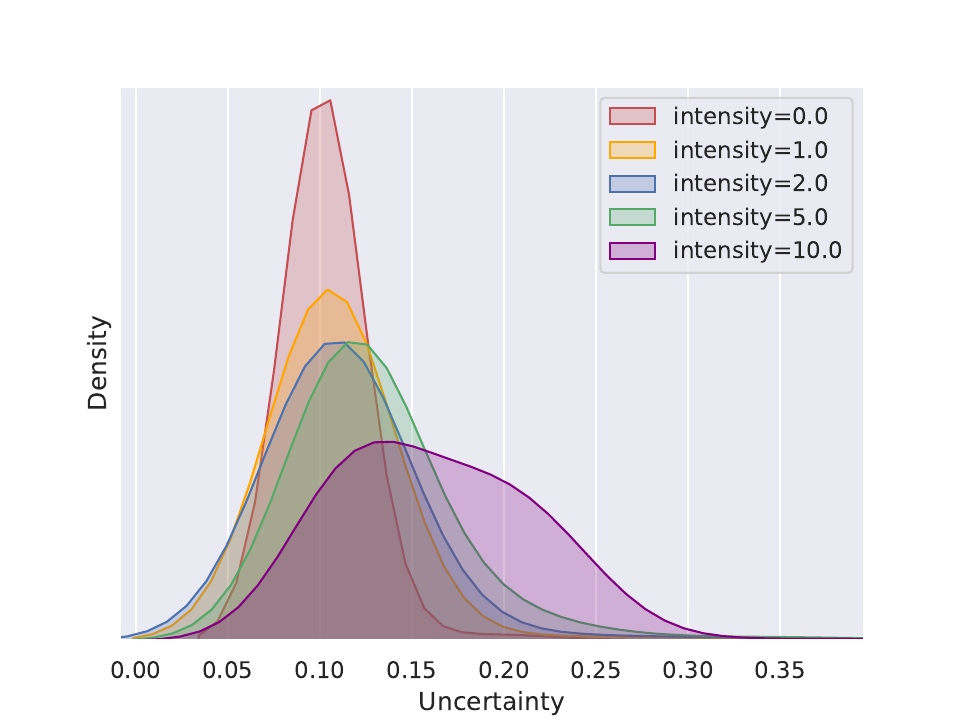}%
		\label{fig_a}}
	\caption{Capturing data uncertainty. The curves of different colors reflect the uncertainty distribution under different noise intensities.}
	\label{capture_uncertainty}
\end{figure*}

To demonstrate that TMSON can capture noise and reflect data uncertainty, a series of experiments are conducted on the CMU-MOSI dataset. Specifically, we follow DUA-Nets~\cite{geng2021uncertainty} to sample noise vectors ($\pmb{\tau}$) from a Gaussian distribution $\mathcal{N}(\pmb{0},\pmb{I})$ and then add these noise vectors multiplied by intensity $\eta$ to contaminate the test data $U$, i.e., $\hat{U}=U+\eta\pmb{\tau}$. Different intensities are used to generate noisy data and observe the uncertainty $\pmb{\sigma}_m$ of each modality. The Gaussian kernel density estimates of uncertainty are depicted in Fig.~\ref{capture_uncertainty}.

There are 3 modalities (text, visual and audio) in the CMU-MOSI dataset, and noise is added to the three modalities simultaneously to verify the impact on the model. The subplots of text and audio modalities show that when the noise intensity is low (intensity is set to 1.0), the distribution curves of noisy and clean data highly overlap, and as the noise intensity increases, the uncertainty increases correspondingly. In addition, the uncertainty distribution of each modality is different, and the text modality is most significantly affected by noise. Furthermore, TMSON efficiently integrates the uncertainty of the unimodal distribution (Fig.~\ref{fig_t}, Fig.~\ref{fig_v}, Fig.~\ref{fig_a}) and reduces the uncertainty of the multimodal distribution (Fig.~\ref{fig_f}). Specifically, when the noise intensity is small, the uncertainty distributions for text, visual and audio modalities are concentrated in the intervals [0, 0.5], [0.1, 0.3] and [0, 0.2], respectively, while the uncertainty of the fused multimodal distribution is concentrated mainly in the [0, 0.3] interval.

\subsubsection{Robustness Evaluation}
\begin{figure}[t]
	\centering
	\includegraphics[width=8cm, height=8cm]{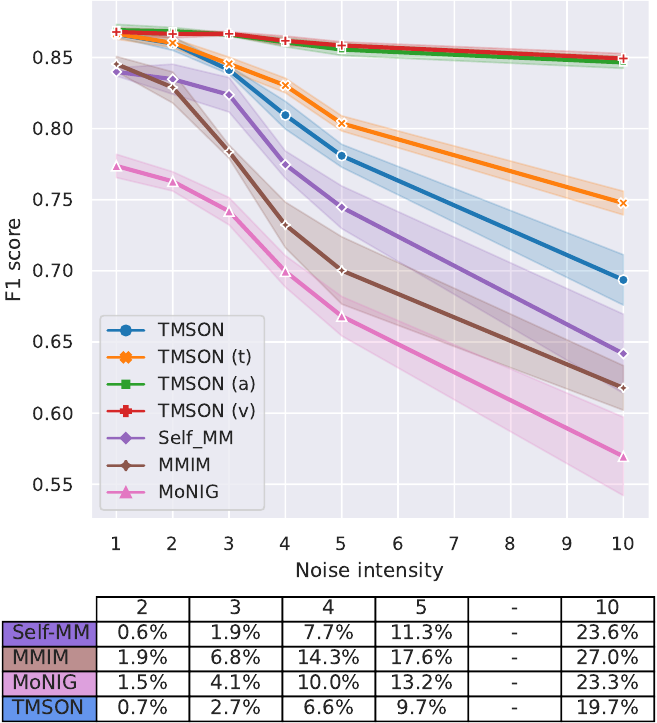}
	\caption{Accuracy with different noise intensities on CMU-MOSI, with tabulated percentage drop for different noise intensities relative to a noise intensity of 1. (t), (v), and (a) are unimodal experiments.}
	\label{noise_intensity}
\end{figure}

\begin{table*}[ht]
	\renewcommand{\arraystretch}{1.2}
	\caption{Performance of each model on CMU-MOSI with different missing rates. Percentages in brackets indicate the percentage drops relative to a missing rate of 0.}
	\label{missing_modality}
	\centering
	\begin{tabular}{c|c| c c c c c }
		\cline{1-7}
		& & \multicolumn{5}{c}{missing rate} \\ 
		\cline{3-7}
		& & 0.0 & 0.1 & 0.2 & 0.3 & 0.5 \\
		\cline{1-7}
		\multirow{2}{*}{Self-MM~\cite{yu2021learning}} 
		& F1  & 85.53 & 81.79 (4.4\%) & 75.41 (12.2\%) & 73.68 (13.9\%) & 62.29 (27.2\%) \\
		& Corr & 0.791 & 0.762 (3.7\%) & 0.705 (10.9\%)  & 0.667 (11.3\%) & 0.534 (32.5\%) \\
		\cline{1-7}
		
		\multirow{2}{*}{MMIM~\cite{han2021improving}} 
		& F1  & 86.10 & 83.08 (3.5\%) & 78.67 (8.6\%) & 76.30 (11.4\%) & 68.96 (19.9\%) \\
		& Corr & 0.797 & 0.756 (5.1\%) & 0.702 (11.9\%)  & 0.654 (17.9\%) & 0.527 (33.9\%) \\
		\cline{1-7}
		
		\multirow{2}{*}{MoNIG~\cite{ma2021trustworthy}} 
		& F1  & 79.00 & 75.07 (5.0\%) & 73.95 (6.4\%) & 71.12 (10.0\%) & 67.58 (14.5\%) \\
		& Corr & 0.633 & 0.581 (8.2\%) & 0.562 (11.2\%) & 0.516 (18.4\%) & 0.454 (28.3\%) \\
		\cline{1-7}
		
		\multirow{2}{*}{TMSON} 
		& F1  & 87.12 & 85.67 (1.7\%) & 81.75 (6.2\%) & 80.05 (8.1\%) & 72.98 (16.2\%) \\
		& Corr & 0.809 & 0.776 (4.1\%) & 0.733 (9.4\%)  & 0.719 (11.1\%) & 0.562 (30.5\%) \\
		\cline{1-7}
	\end{tabular}
\end{table*}

To evaluate the robustness of TMSON, experiments are conducted on CMU-MOSI. For fair comparison, all models are trained on three modalities. These models are then used for noise experiments and modality-missing experiments.
	
In noise experiments, we follow Section~\ref{estimate_uncertainty} and generate test data with different noise intensities. We perform unimodal experiments (add noise to text (t), visual (v), and audio (a) separately for TMSON). Furthermore, multimodal experiments (add noise to three modalities for all models) are also verified. For each model, we perform 5 runs and average the results. The experimental results are depicted in Fig.~\ref{noise_intensity}, and the percentage drop for multimodal experiments is provided in the table. The accuracy of all models generally declines as the noise intensity increases. In  unimodal experiments, visual and audio modalities exhibit minimal changes despite the increased noise intensity, indicating that they are auxiliary modalities with the least impact on sentiment analysis. In multimodal experiments, TMSON outperforms Self-MM, MMIM and MoNIG in terms of performance. As indicated by the percentage drop in the table, TMSON is more robust than all competitors.

In modality-missing experiments, we follow the approach in~\cite{hazarika2022analyzing}, which randomly drops text modalities according to different missing rates (from 0 to 0.5). In Table~\ref{missing_modality}, TMSON outperforms competitors in the two metrics. Especially when the missing rate is increased to 0.3, TMSON still maintains approximately 80\% accuracy on the (non-negative) F1 score. From the drop percentage, TMSON is more stable than Self-MM and MMIM. Moreover, TMSON is superior to MoNIG when the missing rate is lower than 0.3.

\subsubsection{Interpretation of Trusted Multimodalities}
\begin{figure}[t]
	\centering
	\includegraphics[width=8.8cm, height=7cm]{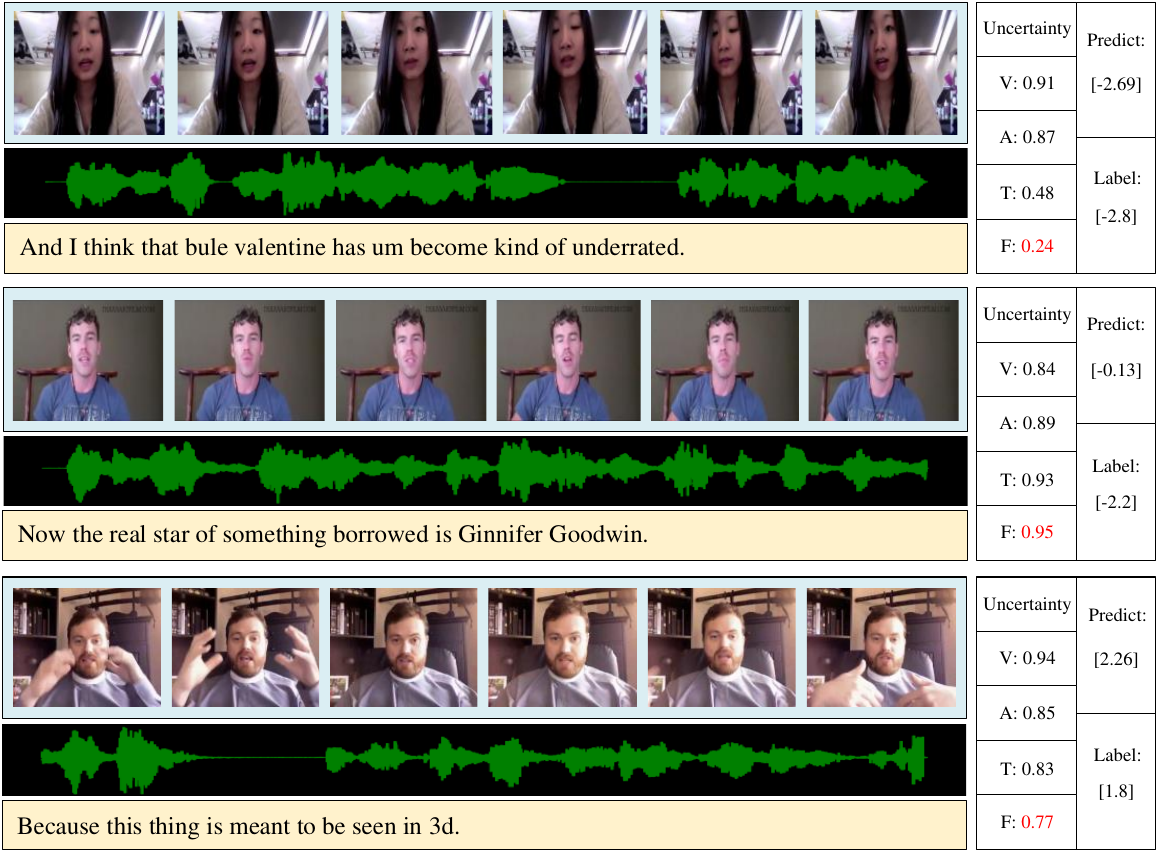}
	\caption{Interpretation of fusion results. ``T", ``V", ``A", and ``F" represent the uncertainty scores for text, visual, audio, and fused modalities, respectively. The predicted sentiment scores and corresponding ground truth labels are also given.}
	\label{video_uncertainty}
\end{figure}
\begin{figure}[t]
	\centering
	\includegraphics[width=8.5cm, height=6.5cm]{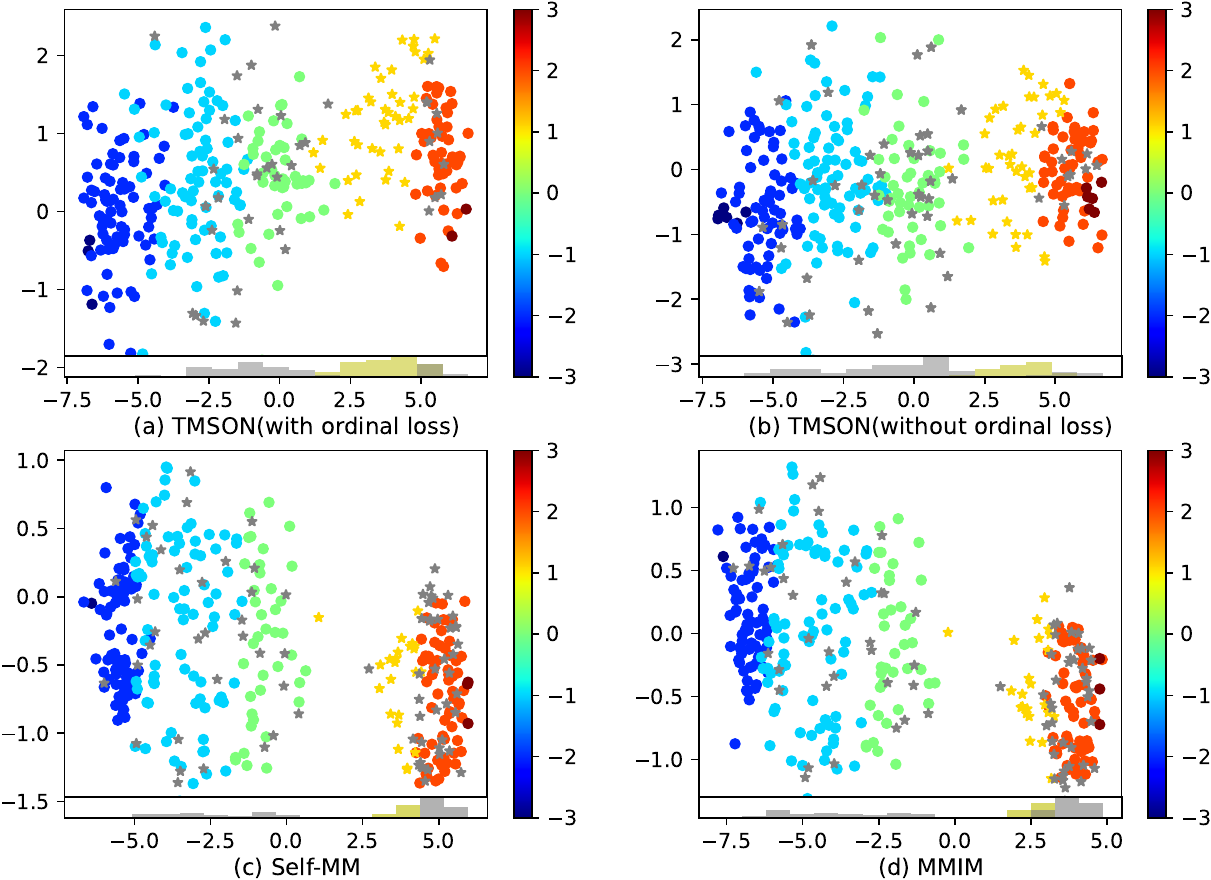}
	\caption{Effect of ordinal regression on CMU-MOSI. Yellow stars are correctly predicted samples, and gray stars are misclassified samples.}
	\label{ordinal_effective}
\end{figure}

An advantage of TMSON is the ability to automatically integrate the uncertainty of multiple modalities to obtain a plausible inference. We approximately explain TMSON's decision-making process by visualizing the uncertainty of the unimodal distribution and fused multimodal distribution. The uncertainty distributions ($\pmb{\mu}_m$ and  $\pmb{\sigma}_m^2$, where $m \in \{f,t,v,a\}$) are vectorized representations, which are converted by the harmonic mean to scalars, ranging from 0 to 1. 

Fig.~\ref{video_uncertainty} shows three cases to illustrate how TMSON works. For the first case, the prediction is correct. The text modality yields lower uncertainty, while the visual and audio modalities provide higher uncertainty. In this case, TMSON considers mainly text modalities and obtains fusion result with a lower uncertainty (F: 0.24). For the second case, the prediction is incorrect. All three modalities provide high uncertainty scores, and the fused result also provides a high uncertainty prediction (F: 0.95), which indicates that the decision is unreliable. For the third case, the prediction is correct. All three modalities provide relatively high uncertainty scores, and the fused result gives a high uncertainty prediction (F: 0.77), which suggests that our prediction is correct (the predicted and ground truth scores all fall into integer 2) but unreliable (i.e., a large gap between the predicted score and ground truth label). Therefore, we truly cannot judge the speaker's attitude from the text modality alone, and further contextual analysis is needed. Experiments emphasize the importance of uncertainty estimation, which provides evidence for our multimodal decision-making.

\subsubsection{Effect of Ordinal Regression}	
Intuitively, an ordinal constraint makes the distribution of samples follow the ordinal relationship, meaning misclassified samples are closer to their actual class distributions. Fig.~\ref{ordinal_effective} visualizes the fused features of models (TMSON, Self-MM and MMIM) via t-SNE~\cite{van2008visualizing}, showing the distribution of misclassified samples, and each subplot provides distribution statistics for misclassified samples. Specifically, we focus on the category with a sentiment score of 1. Fig.~\ref{ordinal_effective}a shows that ordinal regression shifts misclassified samples closer to their actual class. Without ordinal constraints (Fig.~\ref{ordinal_effective}b, Fig.~\ref{ordinal_effective}c and Fig.~\ref{ordinal_effective}d), the distribution of misclassified samples has a larger span.

\subsubsection{Visualize Ordinal Sentiment Space}
\begin{figure}[t]
	\centering
	\includegraphics[width=8.5cm, height=6cm]{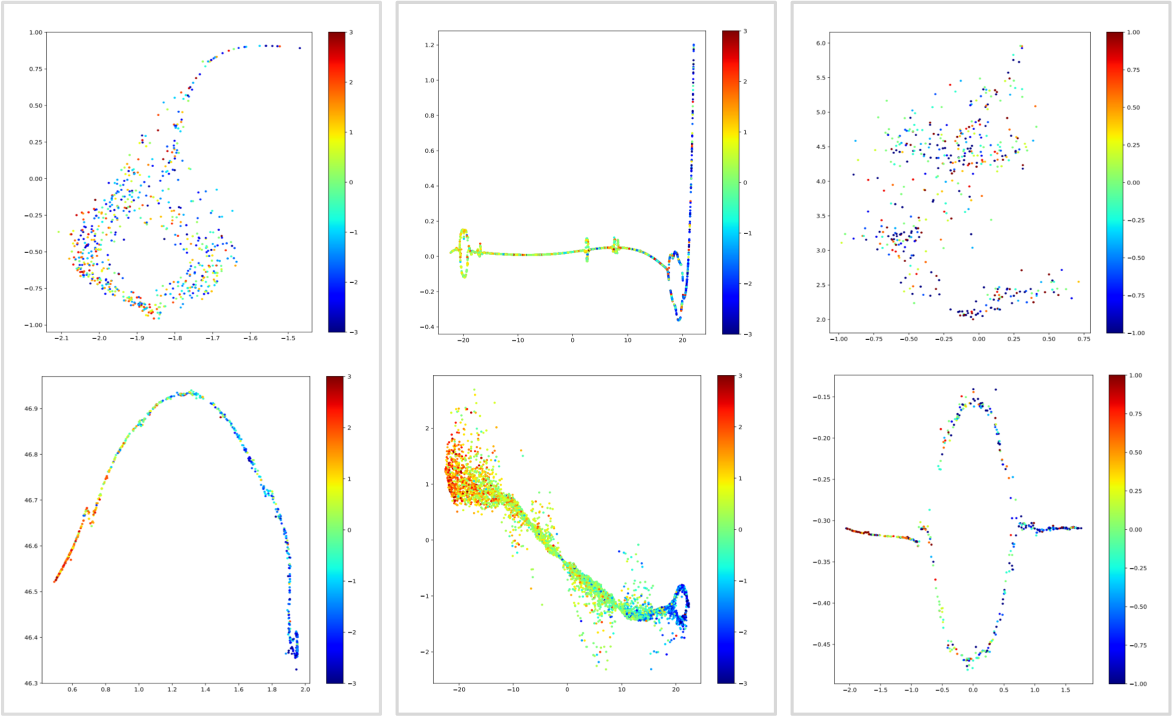}
	\caption{T-SNE visualization of the sentiment space.}
	\label{tsne}
\end{figure}

To demonstrate the ability of TMSON to capture sentiment ordinal relations, we visualize the fused multimodal distribution ($\pmb{\mu}_m$, $\pmb{\sigma}_m^2$) via T-SNE to demonstrate the effectiveness of the ordinal loss. Fig.~\ref{tsne} shows the visualizations, from left to right on the test datasets of CMU-MOSI, CMU-MOSEI and SIMS. Each sample point is assigned a color, with blue for negative and red for positive. The first row has no ordinal loss constraints, while the second row has ordinal constraints. With the ordinal constraint, our model is forced to rank samples of different sentiment categories so that the sentiment space has ordinal properties.

\subsubsection{Complexity Comparison}
\begin{table}[t]
	\renewcommand{\arraystretch}{1.2}
	\caption{Complexity comparison of different methods on CMU-MOSI. ``M" stands for million, and ``s" means seconds.}
	\label{complexity}
	\centering
	\begin{tabular}{c c c c c }
		\toprule
		Method & Parameters & Training/Testing time & Memory  \\
		& (M) & (s) & (GB)  \\
		\midrule
		MAG-BERT~\cite{rahman2020integrating} & 110.71M & 0.048/0.021s & 3.44G\\
		MISA~\cite{hazarika2020misa} & 110.62M & 0.052/0.023s & 3.58G\\
		Self-MM~\cite{yu2021learning} & 109.64M & 0.044/0.025s & 3.57G\\
		MMIM~\cite{han2021improving} & 109.76M & 0.137/0.046s & 4.49G\\
		\midrule
		TMSON & 109.93M & 0.076/0.018s & 3.57G\\
		\bottomrule
	\end{tabular}
\end{table}

We measure the complexity of TMSON by analyzing the number of parameters, training/testing time (per batch) and GPU memory usage. For fair comparison, all experiments are performed on identical hardware configurations. In Table~\ref{complexity}, we compare with Transformer-based methods (MAG-BERT and MISA), as well as state-of-the-art methods (Self-MM and MMIM) on CMU-MOSI. TMSON has fewer parameters than Transformer-based methods. Furthermore, TMSON trains more quickly than MMIM and slightly more slowly than Self-MM. Because TMSON needs to build ordinal relationships in the training phase, while TMSON has a faster performance in the testing phase, where there is no need to build ordinal relationships.

\section{Conclusion}
We propose a framework, TMSON, to adaptively integrate the uncertainty of multiple modalities and obtain consistent trustworthy representations for multimodal sentiment analysis. Specifically, TMSON consists of four modules: unimodal feature representation, uncertain distribution estimation, multimodal distribution fusion, and ordinal sentiment regression. First, the unimodal feature extractor employs observations for each modality, and then the uncertainty estimation module predicts the unimodal distribution over the observations. Then, Bayesian fusion is used on these unimodal distributions, resulting in a multimodal distribution. To introduce ordinal awareness into the sentiment space, ordinal loss is introduced to constrain the multimodal distributions to further improve the sentiment analysis. Extensive experiments demonstrate that TMSON achieves satisfactory performance on multiple datasets and produces reliable sentiment predictions.

\section{Acknowledgment}
This work was supported by the National Natural Science Foundation of China (Nos.61976247).


\end{document}